%% file: PaperForReview.tex
\documentclass[10pt,twocolumn,letterpaper]{article}

\usepackage{wacv}              

\usepackage{graphicx}
\usepackage{amsmath}
\usepackage{amssymb}
\usepackage{booktabs}

\usepackage{listings}
\usepackage{subcaption}
\usepackage{amsmath}
\usepackage{amssymb}
\usepackage{booktabs}
\usepackage{pgfplots}
\usepgfplotslibrary{fillbetween}
\usepackage{tabularx}
\usepackage{framed,multirow}
\usepackage{colortbl} 
\usepackage{tikz}
\usepackage{xcolor}
\usepackage{enumitem}
\usepackage{float}
\usepackage{graphicx}
\usepackage{capt-of}
\usepackage{url}
\usepackage{fancyhdr}
\usepackage{comment}
\usepackage{pgfplotstable} 
\usepackage{graphicx}
\usepackage{booktabs}
\usepackage{colortbl}
\usepackage{pgfplots}
\usepackage{pgfplotstable}
\usepackage[accsupp]{axessibility}  

\usepackage[pagebackref,breaklinks,colorlinks]{hyperref}

\usepackage[accsupp]{axessibility}
\usepackage[capitalize]{cleveref}
\crefname{section}{Sec.}{Secs.}
\Crefname{section}{Section}{Sections}
\Crefname{table}{Table}{Tables}
\crefname{table}{Tab.}{Tabs.}

\def\wacvPaperID{1074} 
\def\confName{WACV}
\def\confYear{2025}

\begin{document}

\title{STLight: a Fully Convolutional Approach for Efficient Predictive Learning by Spatio-Temporal joint Processing}

\author{
    Andrea Alfarano$^{1,2*}$ \quad
    Alberto Alfarano$^{3*}$ \quad
    Linda Friso$^{4}$ \quad
    Andrea Bacciu$^{5\dagger}$ \quad
    \\
    Irene Amerini$^{5}$ \quad
    Fabrizio Silvestri$^{5}$ \\[2pt]
    \small
    $^1$DVS, University of Zurich \quad
    $^2$Max Planck Society \quad
    $^3$Meta \quad
    $^4$Google \quad
    $^5$Sapienza, University of Rome \\
    \tt\small
    andrea.alfarano@uzh.ch, albealfa@meta.com, lfriso@google.com, \\
    \tt\small bacciu@diag.uniroma1.it, amerini@diag.uniroma1.it, fsilvestri@diag.uniroma1.it \\[5pt]
    \small{$^*$Equal contribution \quad $^\dagger$Work done prior to joining Amazon}
}

\maketitle



\input{sec/0_abstract}

\input{sec/1_intro}
\input{sec/2_related_works}
\input{sec/3_methods}

\input{sec/4_experiments}

\input{sec/5_conclusions}

{\small
\bibliographystyle{ieee_fullname}
\bibliography{egbib}
}

\input{sec/x_appendix}

\end{document}

%% file: sec/0_abstract.tex
\begin{abstract}

Spatio-Temporal predictive Learning is a self-supervised learning paradigm that enables models to identify spatial and temporal patterns by predicting future frames based on past frames. Traditional methods, which use recurrent neural networks to capture temporal patterns, have proven their effectiveness but come with high system complexity and computational demand. Convolutions could offer a more efficient alternative but are limited by their characteristic of treating all previous frames equally, resulting in poor temporal characterization, and by their local receptive field, limiting the capacity to capture distant correlations among frames.
In this paper, we propose STLight, a novel method for spatio-temporal learning that relies solely on channel-wise and depth-wise convolutions as learnable layers. STLight overcomes the limitations of traditional convolutional approaches by rearranging spatial and temporal dimensions together, using a single convolution to mix both types of features into a comprehensive spatio-temporal patch representation. This representation is then processed in a purely convolutional framework, capable of focusing simultaneously on the interaction among near and distant patches, and subsequently allowing for efficient reconstruction of the predicted frames. Our architecture achieves state-of-the-art performance on STL benchmarks across different datasets and settings, while significantly improving computational efficiency in terms of parameters and computational FLOPs. The code is publicly available \footnote{\url{https://github.com/AlfaranoAndrea/STLight/}}.

\end{abstract}

%% file: sec/1_intro.tex
\section{Introduction}
\label{sec:intro}

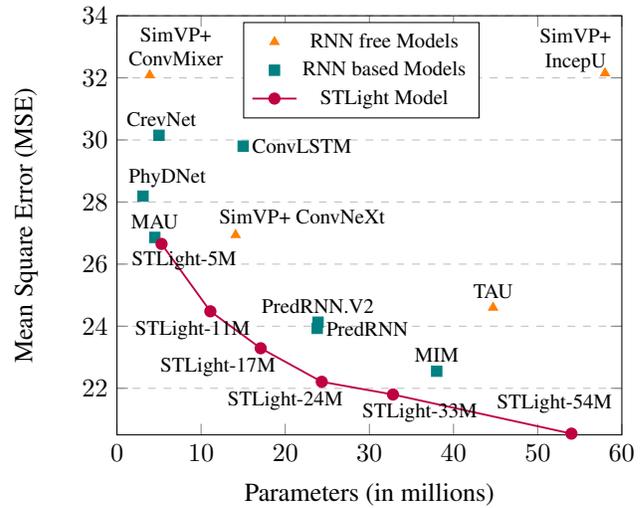
\begin{figure}
 \centering
    \resizebox{0.50\textwidth}{!}{
         \input{plots/MSEvsParams} 
         }
    \caption{MSE vs Number of parameters for existing STL models and STLight on Moving MNIST dataset trained and evaluated under the same settings. }
    \label{fig:msevsparams}
\end{figure}

Spatio-Temporal predictive Learning (STL) aims to extract hidden spatial and temporal patterns by predicting future frames based on previous ones. Utilizing self-supervision, STL models decode complex correlations in sequences of raw data, removing the necessity for labor-intensive manual annotation and facilitating precise forecasts.
STL finds large applications in several resource-constrained domains, such as autonomous driving to predict pedestrian and vehicle movements and prevent accidents \cite{mann2022predicting, akhauri2021improving, cheng2022spatio, gujjar2019classifying}; and in human-robot interactions to anticipate movements and enhance safety \cite{butepage2018anticipating, lee2017learning, zhao2020pose, antonucci2021efficient}. An effective STL model should capture, process, and reconstruct spatial and temporal information from the input frames while balancing efficiency to ensure its applicability in real-world settings.


Relevant methods follow the \textit{Spatial-Temporal-Spatial} framework \cite{tan2023openstl,gao2022simvp, tan2023temporal}, employing a deep Convolutional Neural Network (CNN) to individually encode the spatial details within each frame, a Recurrent Neural Network (RNN) to find and analyze temporal dynamics, and deconvolutional layers to reconstruct frames. However, the sequential nature of RNNs leads to high training and inference costs and limits STL efficiency \cite{gao2022simvp, xu2018predcnn}.


Fully convolutional architectures, which directly predict all the sequences, could be a more efficient alternative to recurrent-based models  \cite{seo2023implicit, xu2018predcnn, tan2023temporal}. However, CNNs are not as effective as RNNs in capturing temporal relationships, as they tend to: (1) treat all frames uniformly, being unable to model the temporal dynamics existing in a continuous Markov process \cite{xu2018predcnn}, and (2) focus on local relationships, thereby losing the global context and necessitating multiple layers to potentially perceive the correlations between distant sequence's portions \cite{cohen2016inductive, wang2024theoretical}.

Attempts to enhance CNNs to surpass their inherent limitations have not succeeded in matching the accuracy of RNNs \cite{tan2023openstl} and may limit model understanding \cite{gao2022simvp}. Investigated enhancements include the introduction of dynamic attention layers \cite{tan2023temporal}, leveraging optical flow information \cite{liu2017video, hu2023dynamic}, incorporating physical knowledge \cite{wu2023pastnet}, proposing tailored losses \cite{tan2023openstl, liu2017video, hu2023dynamic}, and mimicking RNNs' sequential processing \cite{hu2023dynamic, seo2023implicit}.
In our study, we introduce a novel STL architecture that, solely relying on convolutional layers as learnable components, with the focus on minimizing the Mean Square Error (MSE) as the training objective, not only matches or surpasses RNNs in accuracy but also enhances the efficiency of previous CNN models.


We addressed the lack of temporal characterization in CNNs by providing a unified representation of spatial and temporal dimensions and rearranging both dimensions into a comprehensive spatio-temporal patch representation.

Our model further enhances the joint processing of temporal and spatial dynamics, employing a mixer paradigm \cite{chen2021cyclemlp,tolstikhin2021mlp,touvron2022resmlp} which repeatedly integrates inter-patches temporal information and intra-patches spatial information. To overcome CNNs' local receptive field limitations, we explicitly process near and distant intra-patches relationships, inspired by \cite{liu2022more, tan2023temporal}.


Our contributions can be summarized as follows: 
\begin{itemize}
    \item For the first time in the STL context, we jointly process temporal and spatial dynamics, as opposed to the \textit{Spatial-Temporal-Spatial} approach of current methods. In particular, as shown in Fig. \ref{fig:architecture}, we employ (1) a Patch Embedding Encoder for joint spatio-temporal representation with a single convolution layer, (2) a STLMixer with large receptive field as convolutional-only backbone, and (3) an efficient Decoder relying on parameter-free pixel shuffle and a single convolution layer.
    \item We extensively assess our proposed model using challenging STL datasets. We focus on the possibility of scaling efficiently from low-resource to high-accuracy scenarios. Additionally, we rigorously evaluate each component of our model in various settings, establishing a solid foundation for enhanced performance. 
    \item We achieve state-of-the-art results with a full-CNN architecture 
    on major STL datasets, outperforming or matching previous methods in terms of accuracy, parameters, and FLOPs and improved convergence speed during training.

\end{itemize}

%% file: plots/MSEvsParams.tex
\begin{tikzpicture}
\begin{axis}[
    xlabel={Parameters (in millions)},
    ylabel={Mean Square Error (MSE)},
    xmin=0, xmax=60,
    ymin=20.5, ymax=34,
    legend style={at={(0.25,0.87)},anchor=west},
    legend style={font=\footnotesize},
    ymajorgrids=true,
    grid style=dashed,
]
\addplot[scatter src=explicit symbolic, mark=triangle*,only marks,color=orange]
    coordinates {
    (3.9,32.09) 
    (14.1,26.94) 
    (58.0,32.15) 
    (44.7,24.60) 
};

\addplot[scatter src=explicit symbolic, mark=square*, only marks,color=teal, ultra thin,mark options={scale=1}]
    coordinates {
    (15.0,29.80) 
    (12.5,161.38) 
    (3.1,28.19) 
    (23.8, 23.93) 
    (38.0, 22.55) 
    (4.5,26.86) 
    (51.0,35.97) 
    (5.0,30.15) 
    (23.9,24.13) 

};

\addplot[scatter src=explicit symbolic, mark=*,color=purple, mark options={scale=1.0}, line width=0.25mm]
    coordinates {
    (5.3,26.65) 
    (11.10,24.48) 
    (17.08,23.29) 
    (24.34,22.21) 
    (32.8,21.80) 
    (54.01,20.54) 
};

\node[above, align=center, font=\footnotesize] at (axis cs:22,27) {SimVP+ ConvNeXt};
\node[above, align=center, font=\footnotesize] at (axis cs:7,32.09) {SimVP+\\  ConvMixer};
\node[above, align=center, font=\footnotesize] at (axis cs:54.5,31.9) {SimVP+ \\IncepU};
\node[above, font=\footnotesize] at (axis cs:44.7,24.60) {TAU};

\node[right, font=\footnotesize] at (axis cs:15,29.80) {ConvLSTM};
\node[above, font=\footnotesize] at (axis cs:12.5,161.38) {PredNet};
\node[above, font=\footnotesize] at (axis cs:6,28.19) {PhyDNet};
\node[right, font=\footnotesize] at (axis cs:23.8, 23.93) {PredRNN};
\node[above, font=\footnotesize] at (axis cs:38.0, 22.55) {MIM};
\node[above, font=\footnotesize] at (axis cs:4.5,26.86) {MAU};
\node[above, font=\footnotesize] at (axis cs:51.0,35.97) {E3D-LSTM};
\node[above, font=\footnotesize] at (axis cs:5.3,30.15) {CrevNet};
\node[above, font=\footnotesize] at (axis cs:23.9,24.13) {PredRNN.V2};

\node[below, font=\footnotesize] at (axis cs:8,26.7) {STLight-5M};
\node[below, font=\footnotesize] at (axis cs:9,24.48) {STLight-11M};
\node[below, font=\footnotesize] at (axis cs:12,23.29) {STLight-17M};
\node[below, font=\footnotesize] at (axis cs:20,22.21) {STLight-24M};
\node[below, font=\footnotesize] at (axis cs:36, 21.8) {STLight-33M};
\node[above, font=\footnotesize] at (axis cs:52,20.9) {STLight-54M};

\legend{RNN free Models, RNN based Models, STLight Model}
\end{axis}
\end{tikzpicture}

%% file: sec/2_related_works.tex
\section{Background and Related work}

\subsection{Problem definition}
In STL, we aim to forecast future video frames based on past observations. Given the past \(T\) frames up to current time \(t_0\), denoted as
\begin{equation}
    \mathcal{X} = \{x_i\}_{t=t_0-T+1}^{t=t_0} \in \mathbb{R}^{T \times C \times H \times W}
\end{equation}
our objective is to predict the next \(T'\) frames, denoted as
\begin{equation}
    \mathcal{Y} = \{x_i\}_{t=t_0+1}^{t=t_0+T'} \in \mathbb{R}^{T' \times C \times H \times W}
\end{equation}
where each frame \(x_i \in \mathbb{R}^{C \times H \times W}\) is usually a \(H \times W\) image with \(C\) channels.


\subsection{Spatio-Temporal predictive Learning }
Most STL approaches utilize a general \textit{Spatial-Temporal-Spatial} framework \cite{tan2023temporal, gao2022simvp, tan2023openstl}. In this framework, the encoder stage processes the input video frames, focusing on capturing spatial correlations only. The temporal module then leverages the spatial correlations within the frame representation to translate it into a corresponding representation of a future time point. Finally, the decoder stage reconstructs the spatial dimensions of the time-shifted representation, resulting in the output video frames. Based on the type of temporal block used, we can categorize STL models into recurrent-based and recurrent-free models.

\noindent\textbf{Recurrent-based models} 
Recurrent-based models leverage past predictions for individually forecasting each frame, explicitly modeling Markovian temporal evolution but limiting efficiency due to the necessity of making multiple single predictions. In this framework, ConvLSTM \cite{shi2015convolutional} enriches traditional LSTMs \cite{sutskever2014sequence} by integrating convolutional layers, providing both spatial and temporal insights. PredRNN \cite{wang2017predrnn} and its advanced version, PredRNN++ \cite{wang2018predrnn++}, focus on the dual extraction of space and time representations, with the latter improving gradient propagation. MIM \cite{wang2019memory} investigates the video nuances with its self-renewed memory module, capturing the dynamic and stationary aspects of videos. MAU \cite{chang2021mau} utilizes a specialized unit to detect and encapsulate motion patterns within sequential data. IAM4VP \cite{seo2023implicit} and DMVFN \cite{hu2023dynamic} attempt to surpass RNNs limitations by using convolutions, while still retaining recurrent processing.


\noindent\textbf{Recurrent-free models} Recurrent-free approaches improve efficiency by directly forecasting all frames in a sequence, but face challenges with the coherence of predictions, and increased memory consumption as sequence length grows. To improve coherence, DVF \cite{liu2017video} leverages optical flow information, while Pastnet \cite{wu2023pastnet} incorporates physical knowledge. SimVP \cite{gao2022simvp} introduces a seminal framework, which employs stacked convolution layers for both encoding and decoding stages. A UNet architecture \cite{ronneberger2015u} based on inception blocks \cite{szegedy2015going} is placed in the middle as a translator.
Several works have built upon the SimVP framework, focusing particularly on proposing a more competitive translator. TAU \cite{tan2023temporal} incorporates static attention mechanisms to analyze relations between frames and dynamic attention to monitor changes over time. Tan et al. \cite{tan2023openstl} further expand the versatility of the SimVP framework by replacing the temporal translator with successful vision architectures inspired by the transformer model \cite{vaswani2017attention} such as ViT \cite{dosovitskiy2020image}, MLP-Mixer \cite{tolstikhin2021mlp}, ConvMixer \cite{trockman2022patches}, and ConvNeXt \cite{liu2022convnet}, leveraging optical flow information \cite{liu2017video, hu2023dynamic}, incorporating physical knowledge \cite{wu2023pastnet}, and proposing tailored losses \cite{tan2023openstl, liu2017video, hu2023dynamic}.



%% file: sec/3_methods.tex
\section{Methods}
\begin{figure*}[!h]
    \centering
    \includegraphics[width=\textwidth]{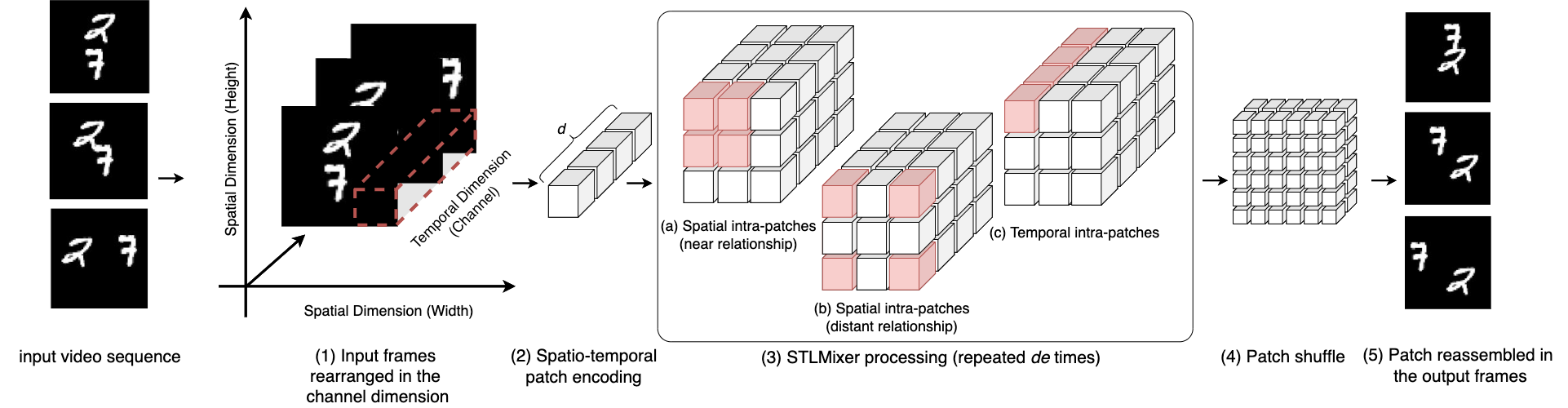}
    \caption{STLight model workflow. We rearrange the input sequence of frames along the channel dimension (1), and through a single convolutional layer, we encode the sequence into patches of size $p \times p$ with hidden temporal dimension $d$, containing both spatial and temporal information (2). The patches are processed through a custom STLMixer block repeated $\texttt{de}$ times (3). Each block processes the relationships between near (a) and distant (b) intra patches along the spatial dimension, as well as the intra-patch relationships on the temporal dimension (c). We decode the output sequence, restoring the initial spatial resolution through a patch shuffle (4) and the temporal resolution by reassembling the patches into the final output sequence (5).}
    \label{fig:architecture}
\end{figure*}

In this section, we present our architecture, depicted in Figure \ref{fig:architecture}. The sequence of input frames is first encoded as spatio-temporal patches (Section \ref{encoder}), processed by STLMixer blocks (Section \ref{STLMixer}), and decoded through patch shuffle and reassemble (Section \ref{decoder}).

\subsection{Spatio-Temporal Patches}\label{encoder}

Given an input batch composed by the sequences of observed frames \(\mathcal{B}_{T} \in \mathbb{R}^{B \times T \times C \times H \times W}\), previous methods \cite{tan2023openstl, gao2022simvp, tan2023temporal} encode each frame individually, rearranging them into the batch dimension and describing the spatial features within each frame of the batch of sequences. Unfortunately, this process treats each frame equally, without explicitly exploiting temporal relationships.
Given the limitations of a fully convolutional approach in capturing both spatial and temporal information, we fuse those dimensions by dividing each frame into patches, with each patch including both spatial and temporal details.

This is achieved by interleaving frames in the channel dimensions \(Z_{T} \in \mathbb{R}^{B \times (T \cdot C) \times H \times W}\), as Voleti et al.~\cite{voleti2022mcvd}, and dividing \(H \times W\) into patches by using a stride of \(p\) and determining the kernel size and padding based on the value of \(O\). Specifically, if \(O \geq 2\), the stride is set to \(p\), the kernel size is \(p \cdot O\), and the padding is \(\left\lfloor \frac{(O - 1) \cdot p}{2}\right\rfloor\). Otherwise, the stride is \(p\), the kernel size is \(p\), and no padding is applied.
To ensure continuity in the representation and maximize spatial information, we adopt extensive patch overlapping. To further enhance efficiency, the encoding is performed with a single convolution.
where \(p\) is the patch size and \(O\) is the desired overlap. Following the convolution, we obtain a tensor \(Z'_{T} \in \mathbb{R}^{B \times d \times H/p \times W/p}\), representing the embedded frames sequence in the hidden dimension \(d\).  

 
Specifically, our embedding strategy aims to preserve the spatial resolution of the embedded tensor as an integer divisor of the initial frames' spatial resolution, allowing the patch shuffle method employed in the decoding phase, as described in Section \ref{sect:reassemble}, and facilitating an efficient finetuning strategy, as discussed in Section \ref{sect:balance}.
To assist the efficient frame representation, we recommend a significant overlap, small patches, and large hidden temporal dimension, as we will demonstrate in the experimental section. This approach minimizes the information compressed per patch while still enabling a \(p^2\) resolution reduction.

\subsection{STLMixer} \label{STLMixer}

After encoding, the input frames' spatio-temporal relations are represented via patches and their connections. To capture the complex dynamics inherent spatio-temporal learning we necessitate to analyze both proximal and distal patch relationships. This is particularly important with small patches, as proposed in Section \ref{encoder}, which increases their number and, consequently, their distances.

To facilitate this processing, we introduce a Mixer architecture \cite{chen2021cyclemlp,tolstikhin2021mlp,touvron2022resmlp}, which cyclically intermixes information among and within patches, respectively in their spatial and temporal dimension. 
To effectively grasp local and global relationships, we propose the STLMixer, an advanced ConvMixer \cite{trockman2022patches} that includes a two-stage intra-patches mixer based on \cite{liu2022more, tan2023openstl}. 
In the STLMixer block, a compact kernel \(k_{T_1}\) captures local fine-grained details. Subsequently, a dilated convolution layer broadens the receptive field to incorporate global information, utilizing a larger kernel size \(k_{T_2}\). This dual convolution approach merges the intricate details of the representation with an overarching contextual understanding. This STLMixer block is repeated $\texttt{de}$ times in our architecture, with a skip connection between the block \(\texttt{de}/3\) and \(2\cdot\texttt{de}/3\),
providing into later stage processing an earlier frames representation to guide output frames reconstruction.

To facilitate better frame representation by encoding each patch with a larger hidden dimension \(d\), as proposed in Section \ref{encoder}, we avoid any intra-patches attention mechanism, due to the quadratic cost associated in \(d\) in a \( d \times d\) attention mechanism. Our approach, which retains a similar depth-wise and point-wise scheme while avoiding any specialized attention mechanisms\cite{tan2023temporal}, contrasts with the assumptions of TAU \cite{tan2023temporal} regarding the efficacy of convolution in processing spatio-temporal dynamics. We investigate the major effectiveness of STLMixer compared to ConvMixer and TAU through an ablation study in Section \ref{sec:ablation}.

\subsection{Patch Shuffle and Reassemble} \label{sect:reassemble}
\label{decoder}
While many methods employ sequences of transposed convolutions to recover lost resolution \cite{gao2022simvp, tan2023openstl, ye2022vptr, tan2023temporal}, our approach utilizes a single convolution coupled with a learnable, parameter-free shuffle technique.

Following our proposed encoder stage (\ref{encoder}), the spatial resolution is diminished by a factor of \(p^2\). 
To efficiently restore the original resolution, we apply the PixelShuffle operator \cite{shi2016real}, which reorganizes elements from a tensor of shape \((B \times d \times H/p \times W/p)\) into a tensor of shape \((B \times d/p^2 \times H \times W)\) achieving the desired output resolution rearranging the patches without learnable layers.

Next, we reassemble the feature space from \((B \times d/p^2 \times H \times W)\) to the targeted \((B \times (T'\cdot C) \times H \times W)\) output sequence. This step is executed through an efficient \(1\times 1\) convolutional layer, reassembling \(d/p^2\) input channels to \(T' \cdot C\) output channels. The process can be represented as follows:
\begin{equation}
    Z'''_{T} = \text{Conv}_{\text{kernel\_size}=1}\left(\text{PixelShuffle}(Z''_{t, T})\right)\end{equation}
With  \(Z''_{T}\) corresponding to the spatio-temporal patches after the STLMixer processing. The final target shape \((B \times T' \times C \times H \times W)\) is restored by reshaping.


%% file: sec/4_experiments.tex
\section{Experiments} \label{experimentsSection}
In this section, we present experiments to validate the effectiveness of our method. Additional Experiments can be found in the Appendix.

\begin{itemize}
    \item \textbf{Standard Spatio-Temporal Predictive Learning:} (Section \ref{sect:standardbenchmarks}) Predicting a constant number of output frames is a standard problem in spatio-temporal predictive learning \cite{tan2023temporal}. We compare STLight with established methods on  Moving MNIST \cite{srivastava2015unsupervised} and TaxiBJ benchmark datasets \cite{zhang2018predicting}.
    
    \item \textbf{Long Sequence Frame Prediction:} (Section \ref{sect:kth}) Predicting longer frame sequences is a central task in STL because it involves a deep understanding of the evolution of the scene. We evaluate our model on the KTH dataset \cite{schuldt2004recognizing} for the task of predicting the next 20 or 40 frames given 10 past observations, also considering the computational demands of processing longer sequences.

    \item \textbf{Self-Supervised Learning Capabilities:} (Section \ref{sec:Unsupervised_effectiveness}) Central to self-supervised learning is the acquisition of robust, domain-independent knowledge by learning more from each data sample. We investigate the domain generalization effectiveness by training STLight models with different parameter ranges (from 0.1 to 15 million) on the KITTI dataset and testing them on the Caltech Pedestrian dataset, assessing efficiency by comparing the training speed of our method with other unsupervised approaches.

    \item \textbf{Hyperparameter Tuning for Accuracy and Efficiency:} (Section \ref{sect:balance}) The impact of STLight's hyperparameter configuration on model performance is explored, offering strategies for fine-tuning these parameters within a specified computational budget to achieve optimal outcomes.

    \item \textbf{Ablation Study:} (Section \ref{sec:ablation}) Through a comprehensive ablation study on various parameter settings, we compare our solutions with conventional practices, providing insights into the efficacy of our methods compared to established approaches.
\end{itemize}

\subsection{Experimental Setups} \label{experimentalSetup}

\noindent \textbf{Datasets} In line with common choices by relevant prior works \cite{wang2018predrnn++, wang2017predrnn, wang2018eidetic}, we quantify the performance of our model on the following synthetic and real-world datasets:
\begin{itemize}
    \item \textbf{Moving MNIST} \cite{srivastava2015unsupervised} (MMNIST) is the fundamental benchmark in STL, consisting of video sequences that depict two independently moving digits. These digits move at different speeds, frequently intersect, and bounce off the edges.
    
    \item \textbf{TaxiBJ} \cite{zhang2018predicting} includes taxi GPS trajectory inflow and outflow data, collected from taxicabs in Beijing. Consistent with prior research \cite{wang2019memory, tan2023temporal}, we normalize the data to the range \([0,1]\).
    
    \item \textbf{KTH} \cite{schuldt2004recognizing} is a human motion dataset that features six types of movements performed by 25 individuals across four different scenarios. In alignment with previous studies \cite{wang2018eidetic, villegas2017decomposing}, we use individuals 1-16 for training and 17-25 for validation.
    
    \item \textbf{KITTI} \cite{Geiger2012CVPR} and \textbf{Caltech} \cite{dollar2009pedestrian} are urban datasets featuring videos from vehicles navigating urban environments. Following the protocols established by previous studies  \cite{lotter2016deep, yu2019crevnet}, we train STLight on the KITTI dataset and evaluate its performance on the Caltech Pedestrian dataset.
\end{itemize}


\begin{table}[!h]

    \centering
    \caption{Datasets composition. The training and testing set have \(N_{\text{train}}\) and \(N_{\text{test}}\) samples, respectively. We predict the future \(T'\) frames from the past $T$.}
    \resizebox{0.5\textwidth}{!}{%
    \begin{tabular}{lccccc}
    \toprule
     & \(N_{\text{train}}\) & \(N_{\text{test}}\) & $(C, H, W)$ & $T$ & $T'$\\
    \midrule
    MMNIST & Generated & 10000 & (1, 64, 64) & 10 & 10 \\
    TaxiBJ & 20461 & 500 & (2, 32, 32) & 4 & 4 \\
    KTH & 4940 & 3030 & (1, 128, 128) & 10 & 20  \\
    Caltech & 2042 & 1983 & (3, 128, 160) & 10 & 1 \\
    \bottomrule
    \end{tabular}
    }
\end{table}
\label{datasetComposition}

\noindent \textbf{Measurement} In this work, we consider both accuracy and computational resources utilization to provide a comprehensive evaluation. Accuracy is measured through Mean Squared Error (MSE), Mean Absolute Error (MAE), Structure Similarity Index Measure (SSIM), and Peak Signal-to-Noise Ratio (PSNR). MSE and MAE estimate the absolute pixel-wise errors, SSIM measures the similarity of structural information within the spatial neighborhoods, and PSNR measures the quality difference between an original and a reconstructed image, quantifying the level of distortion or noise. 
Computational resources are assessed by the parameters count and FLOPs, measured through the fvcore library \cite{fvcore}.

\noindent \textbf{Train-eval settings}
We implement our work using OpenSTL \cite{tan2023openstl}, a well-established open-source framework. Our model is trained using the Mean Squared Error (MSE) as the training objective, and the best hyper-parameters are identified through a grid-search approach. 
To guarantee replicability and fair evaluation, we compare our results against OpenSTL's public results, ensuring the same consistent training setting for all comparisons. 
To robustly evaluate the performance and scalability of STLight, we train multiple instances of the model on each dataset, varying only the number of parameters. The training parameters used for each dataset are reported in Table \ref{tab:hyperparameters}, and we will release OpenSTL configuration files to facilitate the reproduction of our results. Experiments are conducted on a single NVIDIA RTX A6000 GPU with 48GB of VRAM.


\begin{table}[!h]

    \centering
    \caption{Hyperparameters settings for each dataset.}
   \resizebox{0.5\textwidth}{!}{%
\begin{tabular}{lcccc}
\toprule
 & MMNIST & TaxiBJ & KTH & Caltech \\
\midrule
Learning Rate (lr) & 0.003 & 0.003 & 0.0005 & 0.01 \\
Final div factor & 10000& 10000& 10000& 3000 \\
Batch Size & 16 & 16 & 12 & 8 \\
LR Scheduler & OneCycle & Cosine & OneCycle & OneCycle \\
Optimizer & Adam & Adam & Adam & Adam \\
Epochs & 200 & 50 & 100 & 100 \\
\bottomrule
\end{tabular}
}

\label{tab:hyperparameters}
\end{table}

\subsection{Standard STL benchmarks}\label{sect:standardbenchmarks}

\begin{table*}[h!]

\centering
\caption{Quantitative results demonstrating our model's performance in accuracy and computational efficiency compared to OpenSTL's published benchmark baselines under equivalent training and evaluation conditions on MMNIST, TaxiBJ, and KTH datasets.}
\resizebox{\textwidth}{!}{%
\begin{tabular}{l|ccccc|ccccc|ccccc}
\toprule
& \multicolumn{5}{c|}{\textbf{MMNIST}} & \multicolumn{5}{c|}{\textbf{TaxiBJ }} & \multicolumn{5}{c}{\textbf{KTH }} \\
\textbf{Method} & \textbf{MSE} $\downarrow $ & \textbf{MAE} $\downarrow $ & \textbf{SSIM} $\uparrow $  & \textbf{Params} & \textbf{FLOPs} & \textbf{MSE $\times$ 100} $\downarrow $ & \textbf{MAE} $\downarrow $ & \textbf{SSIM} $\uparrow $ & \textbf{Params} & \textbf{FLOPs} & \textbf{MAE} $\downarrow $ & \textbf{PSNR} $\uparrow $  & \textbf{SSIM} $\uparrow $& \textbf{Params} & \textbf{FLOPs} \\
\midrule
\multicolumn{16}{c}{\textbf{Recurrent Methods}} \\
\midrule
ConvLSTM \cite{shi2015convolutional} & 29.80 & 90.64 & 0.9288 & 15.0M & 56.8G  & 33.58 & 15.32 & 0.9836 & 14.98M & 20.74G & 445.5 & 26.99 & 0.8977 & 14.9M & 1368G \\
E3D-LSTM \cite{wang2018eidetic} & 35.97 & 78.28 & 0.9320 & 51.0M & 298.9G & 34.27 & 14.98 & 0.9842 & 50.99M & 98.19G  & 136.40 & 892.7 & 0.8153 & 53.5M & 217G \\
PhyDNet \cite{guen2020disentangling} & 28.19 & 78.64 & 0.9374 & 3.1M & 15.3G & 36.22 & 15.53 & 0.9828 & 3.09M & 5.60G & 765.6 & - & 0.8322 & 3.1M & 93.6G \\
MAU \cite{chang2021mau} & 26.86 & 78.22 & 0.9398 & 4.5M & 17.8G & 32.68 & 15.26 & 0.9834 & 4.41M & 6.02G &471.2 &26.73 & 0.8945 &20.1M &399G \\
MIM \cite{wang2019memory} & 22.55 & 69.97 & 0.9498 & 38.0M & 179.2G & 31.1 & 14.96 & 0.9847 & 37.86M & 64.10G & 380.8 & 27.78 & 0.9025 & 39.8M & 1099G \\
PredRNN \cite{wang2017predrnn} & 23.97 & 72.82 & 0.9462 & 23.8M & 116.0G  & 31.94 & 15.31 & 0.9838 & 23.66M & 42.40G & 380.6 & 27.81 & 0.9097 & 23.6M & 2800G \\
PredRNN++ \cite{wang2018predrnn++} & 22.06 & 69.58 & 0.9509 & 38.6M & 171.7G  & 33.48 & 15.37 & 0.9834 & 38.40M & 62.95G & 370.4 & \textbf{28.13} &\textbf{ 0.9124} & 38.3M & 4162G \\
PredRNNv2 \cite{wang2022predrnn} & 24.13 & 73.73 & 0.9453 & 23.9M & 116.6G & 38.34 & 15.55 & 0.9826 & 23.67M & 42.63G &368.8 & \underline{28.01} & 0.9099 & 23.6M & 2815G \\
DMVFN \cite{hu2023dynamic
} & 123.67 & 179.96 & 0.8140 & 3.5M & 0.2G & 339.5 & 45.526 & 0.8321 & 3.54M & 0.057G &413.2 & 26.65 & 0.8976 & 3.5M &0.88G
 \\
\midrule
\multicolumn{16}{c}{\textbf{Recurrent-free Methods}} \\
\midrule
SimVP+ConvMixer \cite{tan2023openstl} & 32.09 & 88.93 & 0.9259 & 3.9M & 5.5G & 36.34 & 15.63 & 0.9831 & 0.84M & 0.23G  & 446.1 & 26.66 & 0.8993 & 1.5M & 18.3G \\
SimVP+ViT \cite{tan2023openstl} & 35.15 & 95.87 & 0.9139 & 46.1M & 16.9G & - & - & - & - & - & - & - & - & - & - \\
SimVP+InceptU \cite{gao2022simvp} & 32.15 & 89.05 & 0.9402 & 58.0M & 19.4G  & 32.82 & 15.45 & 0.9835 & 13.79M & 3.61G  & 397.1 & 27.46 & 0.9065 & 12.2M & 62.8G \\
TAU \cite{tan2023temporal} & 24.60 & 71.93 & 0.9454 & 44.7M & 16.0G & 31.08 & \textbf{14.93} & \underline{0.9848} & 9.55M & 2.49G & 421.7 & 27.10 & 0.9086 & 15.0M & 73.8G \\
\multicolumn{1}{l|}{\it{STLight-XS  (Ours)}}  & 24.48 & 71.21 & 0.9444 & 11.1M & 10.6G &  - & - & - & - & - & 376.1 &27.50 &0.9052& 1.4M &5.4G \\
\multicolumn{1}{l|}{\it{STLight-S  (Ours)}} & 23.29 & 68.33 & 0.9454 & 17.1M & 16.5G &34.79 &15.58 &0.9825& 0.99M& 0.82G &377.8 & 27.52 &0.9078& 5.4M& 20.9G \\
\multicolumn{1}{l|}{\it{STLight-M  (Ours)}} & \underline{22.21} & \underline{68.33} & \underline{0.9496} & 24.3M & 23.7G & \underline{32.54} & 15.25 & 0.9839 & 1.65M & 1.37G &\underline{367.9} & 27.54 & 0.9102& 9.5M& 37.1G \\

\multicolumn{1}{l|}{\it{STLight-L  (Ours)}} & \textbf{21.80} & \textbf{66.92} &\textbf{ 0.9515} & 32.9M & 32.3G &\textbf{30.87} &\underline{15.00} &\textbf{0.9853} &2.96M& 2.71G &\textbf{363.8} & 27.57 & \underline{0.9113} &14.6M &57.8G \\

\bottomrule
\end{tabular}}
\label{tab:fulltable}
\end{table*}

\subsubsection{Moving MNIST}\label{sect:mmnist}
This dataset is a standard benchmark in STL. We evaluate four variants of STLight, which differ in the number of parameters used, against existing STL methods. The results are reported in Table \ref{tab:fulltable}.

STLight, with its convolutional design, outperforms both recurrent-based and recurrent-free models across all metrics, achieving more accuracy while requiring significantly less computational resources (FLOPs). Remarkably, under standardized settings, STLight-XS surpasses the current state-of-the-art recurrent-free architecture by using only 25\% of its parameters, while STLight-S surpasses the best recurrent architecture with only 14\% of its FLOPs.

In Figure \ref{fig:msevsparams}, we 
demonstrate that STLight can efficiently scale across all ranges and surpass previous methods in both accuracy and efficiency. Qualitative visualizations of the predicted results are shown in Figure \ref{fig:qualitative}. Even when the input frames vary significantly from future frames, our model effectively generates dependable results, capturing movement properties such as direction, speed, and changes in direction at the edges. This underscores the model's capability to capture the underlying dynamics and generate accurate future sequences.

\subsubsection{TaxiBJ}\label{sect:taxibj}

\begin{figure*}[h!]
\centering
    \resizebox{1\textwidth}{!}{
    \centering
    \includegraphics[width=\linewidth]{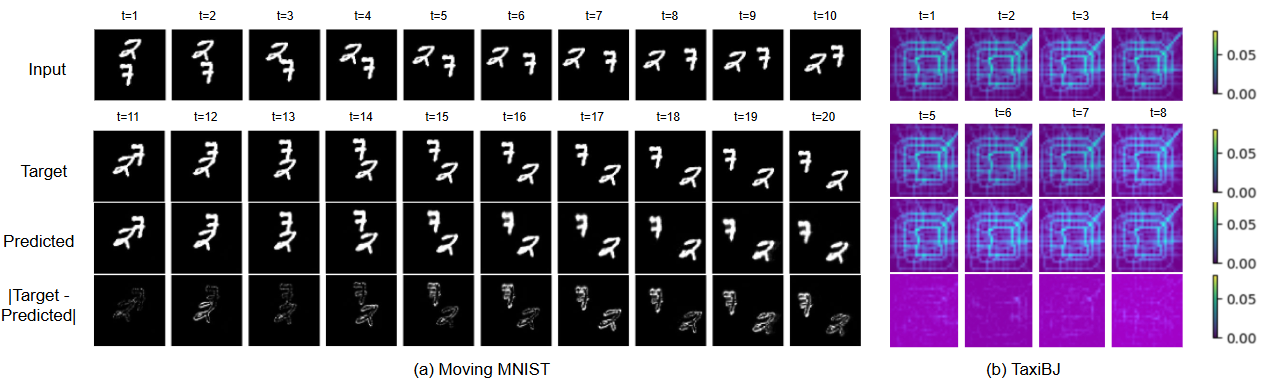}
    }
    \caption{Qualitative results on Moving MNIST and TaxiBJ datasets.
    }
    \label{fig:qualitative}
\end{figure*}

In this section, we evaluate STLight on the TaxiBJ dataset, a standard benchmark in real-world traffic prediction. This dataset poses the challenge of discerning how external factors, from weather shifts to unexpected events, significantly alter traffic behaviors. Given the low-resolution nature of this dataset, which is only \(32 \times 32\) pixels, we utilize a patch size of \(p=1\). As presented in Table \ref{tab:fulltable}, STLight outperforms other notable methods. For example STLight-L achieves the most optimal MSE $\times$ 100 and SSIM, despite having 69\% fewer parameters than the current state-of-the-art model, TAU.
Qualitative results are shown in Figure \ref{fig:qualitative}.

\subsection{Long sequence frames prediction} \label{sect:kth}


Recurrent methods are capable of predicting long coherent sequences of frames from a limited number of past observations, by feeding predicted frames back into the network and recursively outputting predictions. However, recurrent-free methods \cite{gao2022simvp,tan2023temporal} can emulate such processing, but at the cost of efficiency due to the necessity for multiple computations.

In contrast, we evaluate STLight's ability to directly predict the complete target long sequence, leveraging only 10 given input frames to generate the next 20 frames. This approach facilitates efficient parallelization and demonstrates effective long-range video prediction capabilities.

In Table \ref{tab:fulltable} STLight-L and STLight-M match or surpass current state-of-the-art methods. Notably, our STLight-L matches PredRNNv2 while using only 61\% of its parameters and 2\% of its FLOPs. Other STLight variants are even more cost-efficient, with only a slight loss in accuracy. These results highlight STLight's ability to predict extended sequences with high accuracy while requiring significantly fewer computational resources, aligning with findings in Sections \ref{sect:mmnist} and \ref{sect:taxibj}.

\subsection{Unsupervised Learning effectiveness}\label{sec:Unsupervised_effectiveness}
\subsubsection{Domain generalization} \label{sect:adaptation}
Domain adaptation, or out-of-distribution generalization, is a common evaluation scenario for STL\cite{lotter2016deep, gao2022simvp, yu2020efficient}. This evaluation focuses on the challenging task of training a model on one domain, with the goal of successfully generalizing to a new domain. This presents a unique challenge due to differing data distributions across domains, requiring models to extract transferable knowledge and adapt to new settings.

\begin{figure}[h!]
    \centering
       \resizebox{0.45\textwidth}{!}{
     \input{plots/domain}}
    \caption{ STLight models trained on KITTI (0.1M-15M parameters) outperform baselines on Caltech, demonstrating strong cross-dataset generalization with more efficient resource utilization. }
    \label{fig:domain_adaptation}
\end{figure}
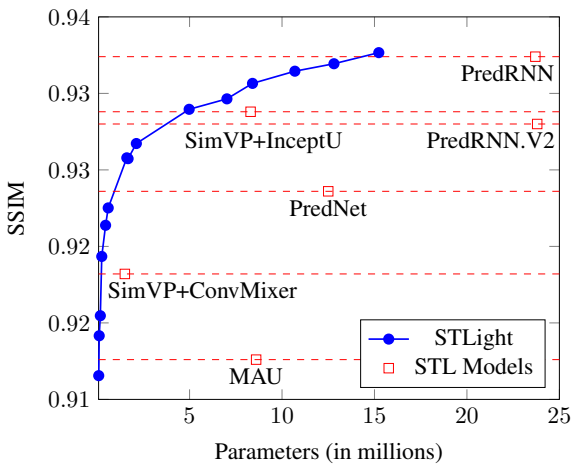

We assess STLight's capability to generalize by using the KITTI and Caltech pedestrian datasets, as described in Section \ref{experimentalSetup}. Additionally, we assess STLight's scaling capabilities by varying the number of its parameters, ranging from 0.1M to 15M, and comparing it with OpenSTL baselines of comparable sizes, not exceeding 25M parameters. Figure \ref{fig:domain_adaptation} presents our findings. STLight consistently demonstrates strong domain generalization across all parameter sets, outperforming notable methods while using significantly fewer parameters. Notably, STLight surpasses competitive models such as MAU, PredRNN.V2, SimVP, and PredRNN, using respectively only 1.5\%, 20\%, 60\%, and 65\% of their parameters.

\subsubsection{Sample Efficiency in Training}\label{sec:trainingEfficiency}
Sample efficiency measures a model's ability to achieve a predefined performance level with the minimal amount of training samples; in other words, it reflects the learning speed from training data \cite{ wang2022sample, weisz2018sample}.

In this section, we train the existing models and STLight-32M on the Moving MNIST dataset under standardized conditions. 

As depicted in Figure \ref{fig:runningTime}, STLight consistently outperforms other methods in terms of sample efficiency. Beyond the initial 21 epochs, STLight achieves a lower MSE compared to established methods across all subsequent training durations. 
This superior performance is exemplified by STLight achieving an MSE of 37.38 at epoch 50, whereas other models require at least an additional 19 epochs to reach comparable accuracy.
This demonstrates STLight's remarkable ability to learn effectively with fewer training samples, highlighting its potential for efficient training and superior data utilization.

\begin{table*}[h!]
\caption{\centering Ablation study of our proposed method}
\resizebox{\textwidth}{!}{%
\begin{tabular}{l|cccc|cccccc}
\toprule
& \multicolumn{4}{c|}{\textbf{Architecture composition}} & \multicolumn{6}{c}{\textbf{MSE for different \# Parameters }} \\
Our modules with ... & Encoder & Decoder & Translator & Inter-block skip connection & 11M & 17M & 24M & 32M & 42M & 79M\\
\toprule

$(B\cdot T)\times C$ Encoder and Decoder & TAU & TAU & Ours & at 1/3 and 2/3 & 27.89 & 26.56 & 25.69 & 25.26 & 24.47 & -\\
TAU Encoder and Decoder & TAU & TAU & Ours & at 1/3 and 2/3 & 27.89 & 26.56 & 25.69 & 25.26 & 24.47 & -\\
Pyramidal convolution Encoder & Pyramid & Ours & Ours & at 1/3 and 2/3 & 26.81 & 25.27 & 23.98 & 23.33 & 22.94 & -\\
Pyramidal deconvolution Decoder & Ours & Pyramid & Ours & at 1/3 and 2/3 & 27.89 & 27.36 & 26.44 & 25.59 & 24.75 & -\\
Standard ConvMixer Translator& Ours & Ours & ConvMixer & NA & 28.64 & 27.81 & 26.53 & 25.74 & 24.82 & -\\
Standard TAU Translator & Ours & Ours & TAU & NA & - & - & - & - & - & 45.12\\
No inter-block skip connection  & Ours & Ours & Ours & no skip & 24.86 & 23.66 & 22.59 & 22.28 & 21.97 & -\\
Skip connection at 1/5 and 4/5   & Ours & Ours & Ours & at 1/5 and 4/5 & 24.94 & 23.68 & 22.62 & 22.13 & 21.92 & -\\ 
\midrule
STLight  (Ours) & Ours & Ours & Ours & at 1/3 and 2/3 & 24.48 & 23.29 & 22.21 & 21.80 & 21.55 & -\\
\bottomrule
\end{tabular}
}

 \label{ablation}
\end{table*}

\begin{figure*}[!h]
\centering 
\begin{minipage}[t]{0.3\textwidth} 
  \centering
   \resizebox{\textwidth}{!}{\input{plots/sample_efficiency}}
\captionof{figure}{Learning curve comparison between state-of-the-art methods and ours.}
\label{fig:runningTime}
\end{minipage}\hfill 
\begin{minipage}[t]{0.3\textwidth} 
\centering
\resizebox{1\textwidth}{!}{\input{plots/dimvsdepth}}
\captionof{figure}{Learning curve comparison for different values of $d$ and $\texttt{de}$.}
\label{fig:tradeoff}
\end{minipage}\hfill
\begin{minipage}[t]{0.3\textwidth} 
\centering
\resizebox{1\textwidth}{!}{\input{plots/overlapping}}
\caption{Learning curves with different patch overlapping, keeping $p = 2$.}
\label{fig:overlappingMSE}
\end{minipage}
\end{figure*}

\subsection{Tuning for accuracy and efficiency}\label{sect:balance}

An effective learning model must balance accuracy with computational efficiency. In this section, we investigate the role of STLMixer's hyperparameters in achieving this balance. We begin by examining their relationship under preliminary settings, then explore strategies for scaling the STLMixer in low-resource environments, and finally, we adjust them for an increase in accuracy, covering a wide spectrum of applications.



\noindent\textbf{Preliminaries}
As shown in Appendix A, the parameter complexity of STLight scales with \(O(\texttt{de} \cdot d^2)\): the balance between the hidden dimension \(d\) of spatio-temporal patches and the number of repeated STLMixer blocks \(\texttt{de}\) is crucial for STLight's optimal efficiency.
We focus on the validation loss curve \(L(\texttt{de}, d)\) (Figure \ref{fig:tradeoff}). In particular, for a fixed \(d\), we are interested in the "elbow" point \cite{ROUSSEEUW198753} of \(L\), which is the point that marks the transition from significant to negligible MSE improvements, thereby providing valuable information regarding the optimal configuration of the model parameters. Empirical data indicates that this elbow occurs when \(\texttt{de} \approx 16\). Additionally, the parameters \(k_{T_1}\) and \(k_{T_2}\) significantly affect STLight's ability to exploit local and global features. As discussed in Appendix B, we find that \(k_{T_1} = 3\) and \(k_{T_2} = 7\) are optimal for general applications.


\noindent\textbf{Low parameters tuning}
To effectively reduce the STLight parameter count and limit the loss in accuracy, we maintain fixed to their optimal values (\(k_{T_1} = 3\), \(k_{T_2} = 7\) and \(\texttt{de} = 16\)), and we tune the parameter \(d\),  scaling the parameters count by a \(d^2\) factor, considering that STLMixer blocks has complexity \(O(\texttt{de} \cdot d^2)\), as described in Appendix A. Reducing \(d\) limits the expressive power of each patch. Hence we suggest to represent the input sequences by more patches, reducing their size dimension \(p\). Moreover, with a reduced \(d\), the increase in FLOPs caused by more patches is mitigated, achieving an advantageous balance.

\noindent \textbf{High accuracy tuning}
To increase STLight accuracy, efficiently utilizing more computational budget, we propose to maintain fixed to their optimal values \(k_{T_1}\), \(k_{T_2}\) and \(\texttt{de}\), and we suggest to improve patch representation power, increasing  \(d\) and using large patch overlapping. In Figure \ref{fig:overlappingMSE}, we explore the impact of the overlapping parameter \(O\) on the validation loss across different model sizes. Our empirical analysis demonstrates that larger models benefit from large overlap (\(O=p=2\)), while smaller models achieve optimal performance with no overlap (\(O=0\)).

\subsection{Ablation study}\label{sec:ablation}
We assess our methods against well-established practices in STL through an extensive ablation study. Our evaluations, summarized in Table \ref{ablation}, involve replacing each of our components with a comparable architecture, while maintaining a parameter range between 11M and 42M and ensuring a uniform depth of 16 layers.

We validate the integration of spatial and temporal data into spatio-temporal patches from two perspectives: by replacing the encoder and decoder within the SimVP framework \cite{gao2022simvp,tan2023temporal}, and by comparing our approach of using a single convolutional layer against multiple convolutions in encoders or decoders. This replacement led to decreased performance across configurations due to the increased computational demands, despite maintaining the translator block's budget. Additionally, we assessed our mixer backbone by incorporating the STLMixer with a spatio-temporal layer from TAU \cite{tan2023temporal}. We observed that adding extra attention layers significantly increased the parameter count leading to inefficiency. These findings suggest the potential for exploring attention layers that are better suited for spatio-temporal patches. Substituting the translator with ConvMixer \cite{trockman2022patches} underscored the importance of a wide receptive field in STL. Modifying or removing the skip connection led to performance drops, indicating a crucial balance to be found between the 1/3 and 2/3 layers for efficient information flow.

%% file: plots/domain.tex
\begin{tikzpicture}
\begin{axis}[
    xlabel={Parameters (in millions)},
    ylabel={SSIM},
    xmin=0.08, xmax=25,  
    ymin=0.915, ymax=0.94,
    grid style=dashed,
    legend pos=south east,
    legend columns=1,
]

\addplot[scatter src=explicit symbolic, color=blue,mark=*, line width=0.25mm] coordinates {
    (15.233, 0.937663) 
    (12.809, 0.936939) 
    (10.695, 0.936451)
    (8.404,  0.935654)
    (7.02,   0.934643)
    (4.978,  0.93396)
    (2.122,  0.931723)
    (1.675,  0.930734)
    (1.606,  0.930788)
    (0.6,    0.927515)
    (0.458,  0.926379)
    (0.251,  0.924343)
    (0.174,  0.920469)
    (0.113,  0.91916)
    (0.087,  0.916552) 
    
};
\addplot[scatter src=explicit symbolic, only marks,  color=red,mark=square]
    coordinates {
    (23.7, 0.9374) 
    (8.3, 0.9338) 
    (23.8, 0.9330) 
    (12.5, 0.9286) 
    (1.5, 0.9232) 
    (8.6, 0.9176) 

};

\node[below] at (axis cs:22.2, 0.9374) {PredRNN};
\node[below] at (axis cs:9, 0.9330) {SimVP+InceptU};
\node[below] at (axis cs:21.3, 0.9330) {PredRNN.V2};
\node[below] at (axis cs:12.5, 0.9286) {PredNet};
\node[below] at (axis cs:5.7, 0.9232) { SimVP+ConvMixer};
\node[below] at (axis cs:8.6, 0.9176) {MAU};

\draw[dashed,red] (axis cs:0.08,0.9374) -- (axis cs:30,0.9374) ;
\draw[dashed,red] (axis cs:0.08,0.9338) -- (axis cs:30,0.9338) ;
\draw[dashed,red] (axis cs:0.08,0.9330) -- (axis cs:30,0.9330) ;
\draw[dashed,red] (axis cs:0.08,0.9286) -- (axis cs:30,0.9286) ;
\draw[dashed,red] (axis cs:0.08,0.9232) -- (axis cs:30,0.9232) ;
\draw[dashed,red] (axis cs:0.08,0.9176) -- (axis cs:30,0.9176) ;


\legend{STLight, STL Models}
\end{axis}
\end{tikzpicture}

%% file: plots/sample_efficiency.tex
   \begin{tikzpicture}

\definecolor{clr1}{RGB}{255,0,0}
\definecolor{clr2}{RGB}{0,255,0}
\definecolor{clr3}{RGB}{0,179,0}
\definecolor{clr4}{RGB}{0,102,0}
\definecolor{clr5}{RGB}{0,0,255}
\definecolor{clr6}{RGB}{0,0,204}
\definecolor{clr7}{RGB}{0,0,153}
\definecolor{clr8}{RGB}{0,0,102}

\begin{axis}[
    title={},
    xlabel={Epochs},
    ylabel={MSE},
    xmin=0, xmax=200,
    ymin=20, ymax=160,
    ymode=log,
    log basis y=2,
    legend pos=north east,
    legend style={font=\small},
    ymajorgrids=true,
    legend columns=2,
    grid style=dashed,
]
\addplot[clr1,mark='.',smooth, line width=0.25mm] coordinates {(0,128.3583984375) [0](1,113.863700866699) [1](2,106.324386596679) [2](3,99.1720657348632) [3](4,93.7358856201171) [4](5,88.7590026855468) [5](6,85.270393371582) [6](7,82.7266616821289) [7](8,79.6833724975586) [8](9,75.9734725952148) [9](10,75.596565246582) [10](11,73.3959503173828) [11](12,72.9584274291992) [12](13,68.9300765991211) [13](14,67.0477981567382) [14](15,66.5641784667968) [15](16,64.1084213256836) [16](17,63.3222427368164) [17](18,60.6264762878418) [18](19,59.952781677246) [19](20,57.6967811584472) [20](21,58.0917358398437) [21](22,56.1213569641113) [22](23,56.1660766601562) [23](24,53.5192375183105) [24](25,52.1236381530761) [25](26,53.353214263916) [26](27,51.2441329956054) [27](28,50.3735122680664) [28](29,49.2420768737793) [29](30,50.8077049255371) [30](31,50.5088005065918) [31](32,47.5680389404296) [32](33,46.1010131835937) [33](34,46.1134986877441) [34](35,46.3474998474121) [35](36,44.7675628662109) [36](37,44.101219177246) [37](38,45.4713096618652) [38](39,45.5121192932128) [39](40,43.4022178649902) [40](41,43.1802520751953) [41](42,43.3186416625976) [42](43,42.0406112670898) [43](44,40.8076820373535) [44](45,42.4054069519043) [45](46,41.0927124023437) [46](47,40.223030090332) [47](48,40.0579452514648) [48](49,40.3628997802734) [49](50,41.6208584518432) [50](51,39.991917892456) [51](52,39.9918988418579) [52](53,40.562310546875) [53](54,39.1092008285522) [54](55,39.0987447662353) [55](56,38.2336515579223) [56](57,37.6776976013183) [57](58,39.1669172897338) [58](59,37.4547896270752) [59](60,37.3808035964966) [60](61,36.5794997558593) [61](62,35.8905044174194) [62](63,36.9202361907958) [63](64,35.5163322525024) [64](65,35.6310058898925) [65](66,35.1636063156128) [66](67,34.9652458953856) [67](68,35.0172898330688) [68](69,34.3158013610839) [69](70,33.6692674179077) [70](71,34.7481935272216) [71](72,34.5590761184692) [72](73,33.6932543029784) [73](74,34.2922647933959) [74](75,33.6724032135009) [75](76,33.8226183853149) [76](77,32.2433108520507) [77](78,32.1774577560425) [78](79,32.0665950469971) [79](80,32.4932041091918) [80](81,32.5315151367187) [81](82,31.6723518753051) [82](83,32.0329037322998) [83](84,31.598104560852) [84](85,31.8405400390625) [85](86,31.2570020065307) [86](87,31.1208631134032) [87](88,30.8869859542846) [88](89,30.6090681152343) [89](90,30.5274412460327) [90](91,30.4991838989258) [91](92,30.4099219131469) [92](93,31.1645739593505) [93](94,29.6424010543823) [94](95,29.7222178955078) [95](96,29.5529025802612) [96](97,29.9431289672851) [97](98,29.2458175888061) [98](99,29.1684288482665) [99](100,29.2093929977416) [100](101,28.9647259674072) [101](102,29.1616749954223) [102](103,29.0555100860596) [103](104,28.5017572097777) [104](105,28.4470067901611) [105](106,28.3527990493774) [106](107,28.1263395690918) [107](108,28.2606760635376) [108](109,28.2032744293213) [109](110,28.0527445907592) [110](111,27.4906279907226) [111](112,27.8809936141967) [112](113,27.5552985992431) [113](114,27.1939533309936) [114](115,27.1709583587646) [115](116,27.2673691329955) [116](117,27.1103352355956) [117](118,27.6205033340453) [118](119,27.2984720183752) [119](120,26.8839940338134) [120](121,27.5322312850952) [121](122,26.4120511779785) [122](123,26.6894483566283) [123](124,26.6590509643554) [124](125,26.8044634323119) [125](126,26.2306585998535) [126](127,26.0233012161255) [127](128,25.7629595947265) [128](129,26.0520221176146) [129](130,26.476311843872) [130](131,25.4636165542602) [131](132,25.3413440856933) [132](133,25.544669380188) [133](134,25.2102978363037) [134](135,25.8656132583617) [135](136,25.1844113464355) [136](137,24.7980054244994) [137](138,24.7989495849609) [138](139,26.0112527694702) [139](140,24.9579996185303) [140](141,24.7016710586547) [141](142,25.0573153076171) [142](143,24.7281832504272) [143](144,24.5384116210937) [144](145,24.9876761703491) [145](146,24.3410621185303) [146](147,24.9436107406615) [147](148,23.7028366088866) [148](149,24.1257991790771) [149](150,23.8483905792236) [150](151,23.7198619842529) [151](152,23.5872402191162) [152](153,23.5959758758544) [153](154,23.3800525665283) [154](155,23.4394149780273) [155](156,23.2447662353515) [156](157,23.3430976867675) [157](158,23.3065738677978) [158](159,23.4883346557617) [159](160,23.1171779632568) [160](161,23.0582122802734) [161](162,22.8747196197509) [162](163,22.8196563720703) [163](164,22.677490234375) [164](165,22.7171840667724) [165](166,22.6506633758544) [166](167,22.5965633392334) [167](168,22.5640621185302) [168](169,22.4783496856689) [169](170,22.4179897308349) [170](171,22.3488082885742) [171](172,22.2780055999755) [172](173,22.5224094390869) [173](174,22.1895446777343) [174](175,22.1971092224121) [175](176,22.1232711791992) [176](177,22.122622680664) [177](178,22.0697242736816) [178](179,22.0029880523681) [179](180,22.040612411499) [180](181,22.057530593872) [181](182,21.9823619842529) [182](183,21.9354660034179) [183](184,21.8916065216064) [184](185,21.8663742065429) [185](186,21.8408100128173) [186](187,21.8203899383544) [187](188,21.8334381103515) [188](189,21.8088352203369) [189](190,21.5389091491699) [190](191,21.7567913055419) [191](192,21.7644493103027) [192](193,21.5107852935791) [193](194,21.7632648468017) [194](195,21.7667896270751) [195](196,21.733203125) [196](197,21.5044490814209) [197](198,21.4977809906005) [198](199,21.7661354064941) [199](200,21.4960472106933) [200]};


\addplot[color=clr5,mark='.',smooth, line width=0.25mm] coordinates {(0,120.270637512207) [1200](1,111.283477783203) [1201](2,105.753662109375) [1202](3,100.003540039062) [1203](4,96.3393249511718) [1204](5,93.3004760742187) [1205](6,93.3924331665039) [1206](7,88.9235687255859) [1207](8,86.8255157470703) [1208](9,85.4507369995117) [1209](10,84.0252151489257) [1210](11,81.8796691894531) [1211](12,80.9991302490234) [1212](13,79.7632141113281) [1213](14,79.3789138793945) [1214](15,78.4040069580078) [1215](16,77.9685592651367) [1216](17,75.9189682006836) [1217](18,75.3399200439453) [1218](19,74.5180892944336) [1219](20,73.4984283447265) [1220](21,73.1151733398437) [1221](22,72.763931274414) [1222](23,70.8530883789062) [1223](24,70.3934860229492) [1224](25,71.6619644165039) [1225](26,68.7938919067382) [1226](27,69.8013916015625) [1227](28,69.1307601928711) [1228](29,65.3805313110351) [1229](30,66.5335311889648) [1230](31,65.2547302246093) [1231](32,63.6205673217773) [1232](33,65.5410614013671) [1233](34,63.4262313842773) [1234](35,62.8143844604492) [1235](36,62.8687438964843) [1236](37,63.2352256774902) [1237](38,60.9685287475585) [1238](39,62.0025482177734) [1239](40,61.0680999755859) [1240](41,60.6069297790527) [1241](42,58.1347122192382) [1242](43,57.9366912841796) [1243](44,59.0761566162109) [1244](45,56.8387985229492) [1245](46,56.8074493408203) [1246](47,57.2546005249023) [1247](48,55.6054992675781) [1248](49,56.0925827026367) [1249](50,55.7750930786132) [1250](51,56.0724334716796) [1251](52,55.1503944396972) [1252](53,54.5944900512695) [1253](54,52.9808845520019) [1254](55,53.4143295288085) [1255](56,53.3281669616699) [1256](57,52.9464416503906) [1257](58,51.9876899719238) [1258](59,51.6621398925781) [1259](60,51.4701194763183) [1260](61,51.3048439025878) [1261](62,51.0820846557617) [1262](63,50.5182723999023) [1263](64,49.6449890136718) [1264](65,49.0533599853515) [1265](66,49.6302413940429) [1266](67,48.7474060058593) [1267](68,49.8436851501464) [1268](69,48.5359153747558) [1269](70,48.1719093322753) [1270](71,47.2162322998046) [1271](72,48.7149810791015) [1272](73,46.1041717529296) [1273](74,47.0081100463867) [1274](75,46.9567718505859) [1275](76,47.049819946289) [1276](77,45.9615249633789) [1277](78,46.325569152832) [1278](79,46.4014587402343) [1279](80,45.3230590820312) [1280](81,45.4507751464843) [1281](82,45.4762535095214) [1282](83,44.9771728515625) [1283](84,44.3811264038085) [1284](85,44.6202239990234) [1285](86,45.0585708618164) [1286](87,44.001853942871) [1287](88,43.9142150878906) [1288](89,43.6065635681152) [1289](90,43.5382385253906) [1290](91,42.8318481445312) [1291](92,43.4700393676757) [1292](93,42.6092681884765) [1293](94,42.6392517089843) [1294](95,43.1910247802734) [1295](96,42.6368789672851) [1296](97,41.6559753417968) [1297](98,42.1756057739257) [1298](99,41.9397659301757) [1299](100,41.5371856689453) [1300](101,41.7129135131835) [1301](102,41.7611122131347) [1302](103,41.1477470397949) [1303](104,41.4999847412109) [1304](105,41.1186218261718) [1305](106,41.5147857666015) [1306](107,40.5840301513671) [1307](108,40.2148933410644) [1308](109,40.221019744873) [1309](110,40.0892868041992) [1310](111,40.1038208007812) [1311](112,39.624225616455) [1312](113,39.8254852294921) [1313](114,39.5132865905761) [1314](115,39.3732376098632) [1315](116,38.7617950439453) [1316](117,39.3048820495605) [1317](118,39.7661437988281) [1318](119,38.8388938903808) [1319](120,39.0684661865234) [1320](121,38.6556091308593) [1321](122,38.2823638916015) [1322](123,38.4456901550293) [1323](124,37.9275588989257) [1324](125,38.6707534790039) [1325](126,38.3760757446289) [1326](127,37.7340393066406) [1327](128,37.4991111755371) [1328](129,37.9047317504882) [1329](130,37.9063339233398) [1330](131,37.29736328125) [1331](132,37.1018447875976) [1332](133,37.1457748413085) [1333](134,36.8633804321289) [1334](135,36.4970779418945) [1335](136,36.6492309570312) [1336](137,36.7106628417968) [1337](138,36.8695373535156) [1338](139,36.5479545593261) [1339](140,36.5810012817382) [1340](141,35.5854110717773) [1341](142,36.0922813415527) [1342](143,35.9522857666015) [1343](144,35.5898513793945) [1344](145,35.4375839233398) [1345](146,35.3343429565429) [1346](147,35.600242614746) [1347](148,35.1864356994628) [1348](149,35.2692108154296) [1349](150,34.6447296142578) [1350](151,34.8957252502441) [1351](152,34.7069549560546) [1352](153,34.5172653198242) [1353](154,34.8645629882812) [1354](155,34.4129447937011) [1355](156,34.9146270751953) [1356](157,34.5215835571289) [1357](158,34.584617614746) [1358](159,34.2817955017089) [1359](160,34.1937408447265) [1360](161,34.119773864746) [1361](162,34.2848205566406) [1362](163,34.1932296752929) [1363](164,33.7661437988281) [1364](165,33.7915992736816) [1365](166,33.5468978881835) [1366](167,33.5681533813476) [1367](168,33.3809051513671) [1368](169,33.2943496704101) [1369](170,33.4305038452148) [1370](171,33.3996696472168) [1371](172,33.2752838134765) [1372](173,33.3798675537109) [1373](174,33.1707687377929) [1374](175,32.9425048828125) [1375](176,33.012222290039) [1376](177,32.9433555603027) [1377](178,32.9448699951171) [1378](179,32.8369140625) [1379](180,32.8930892944335) [1380](181,32.9489822387695) [1381](182,32.8613662719726) [1382](183,32.7531051635742) [1383](184,32.5744514465332) [1384](185,32.7105102539062) [1385](186,32.6818237304687) [1386](187,32.6139793395996) [1387](188,32.6098670959472) [1388](189,32.5695266723632) [1389](190,32.6096000671386) [1390](191,32.5987548828125) [1391](192,32.5778007507324) [1392](193,32.5487670898437) [1393](194,32.5718994140625) [1394](195,32.5716934204101) [1395](196,32.5591735839843) [1396](197,32.5483093261718) [1397](198,32.5479736328125) [1398](199,32.5470809936523) [1399](200,32.1478042602539) [1400]};

\addplot[clr2,mark='.',smooth, line width=0.25mm] coordinates {(0,144.682678222656) [300](1,164.786743164062) [301](2,142.470428466796) [302](3,130.095291137695) [303](4,113.098152160644) [304](5,113.225463867187) [305](6,107.090393066406) [306](7,103.643447875976) [307](8,97.6734313964843) [308](9,95.8223419189453) [309](10,101.000259399414) [310](11,91.9658660888671) [311](12,89.9131469726562) [312](13,96.0297470092773) [313](14,90.1686096191406) [314](15,96.1888122558593) [315](16,99.4721221923828) [316](17,76.5952453613281) [317](18,87.2729339599609) [318](19,79.1348876953125) [319](20,79.62109375) [320](21,79.6186370849609) [321](22,79.6860046386718) [322](23,78.7200393676757) [323](24,78.9450302124023) [324](25,79.637954711914) [325](26,68.924072265625) [326](27,71.8441772460937) [327](28,78.1117630004882) [328](29,70.7455520629882) [329](30,75.2375869750976) [330](31,70.279052734375) [331](32,72.0321502685546) [332](33,85.8420257568359) [333](34,70.3474197387695) [334](35,69.6952667236328) [335](36,70.6455993652343) [336](37,69.2930603027343) [337](38,69.6389465332031) [338](39,64.839111328125) [339](40,61.7870483398437) [340](41,63.2373275756835) [341](42,64.0225067138671) [342](43,63.5863037109375) [343](44,62.17822265625) [344](45,58.0036888122558) [345](46,60.2393760681152) [346](47,58.7875518798828) [347](48,57.3279495239257) [348](49,54.3959426879882) [349](50,56.260440826416) [350](51,55.2635307312011) [351](52,53.2169647216796) [352](53,54.8441772460937) [353](54,56.9056091308593) [354](55,54.1199569702148) [355](56,52.2648162841796) [356](57,53.0908279418945) [357](58,53.2500762939453) [358](59,52.7544860839843) [359](60,50.8658142089843) [360](61,50.5616607666015) [361](62,52.4771881103515) [362](63,51.2803306579589) [363](64,49.7025909423828) [364](65,51.9993057250976) [365](66,49.6917343139648) [366](67,48.0133399963378) [367](68,48.582534790039) [368](69,47.0574264526367) [369](70,47.1535034179687) [370](71,47.445686340332) [371](72,46.1528358459472) [372](73,48.5409469604492) [373](74,46.4175834655761) [374](75,48.1206474304199) [375](76,45.4618682861328) [376](77,46.4749069213867) [377](78,46.0352249145507) [378](79,46.2528190612793) [379](80,45.6370849609375) [380](81,45.8916358947753) [381](82,44.7360229492187) [382](83,45.7675056457519) [383](84,45.2775421142578) [384](85,44.8967018127441) [385](86,44.8232803344726) [386](87,43.6986770629882) [387](88,43.5649337768554) [388](89,43.8143081665039) [389](90,43.16743850708) [390](91,43.4098320007324) [391](92,43.7180557250976) [392](93,43.201847076416) [393](94,44.1154708862304) [394](95,42.7298355102539) [395](96,41.7503814697265) [396](97,41.3827743530273) [397](98,42.1028671264648) [398](99,41.684326171875) [399](100,42.2144317626953) [400](101,41.774185180664) [401](102,41.5986900329589) [402](103,42.5075035095214) [403](104,41.8558845520019) [404](105,40.6584854125976) [405](106,40.3036422729492) [406](107,40.8776359558105) [407](108,39.7473526000976) [408](109,39.7415237426757) [409](110,39.5290451049804) [410](111,39.8953704833984) [411](112,39.5914573669433) [412](113,38.7506980895996) [413](114,38.132080078125) [414](115,38.6789398193359) [415](116,38.7021789550781) [416](117,38.704345703125) [417](118,38.6302185058593) [418](119,37.7779655456543) [419](120,37.7072067260742) [420](121,37.4696922302246) [421](122,37.3942527770996) [422](123,38.3255767822265) [423](124,37.5983695983886) [424](125,37.3420181274414) [425](126,36.961368560791) [426](127,37.9318351745605) [427](128,36.15869140625) [428](129,36.9223442077636) [429](130,36.2380065917968) [430](131,36.3757896423339) [431](132,36.1809997558593) [432](133,35.8761291503906) [433](134,35.402229309082) [434](135,35.707649230957) [435](136,35.3236961364746) [436](137,35.5080413818359) [437](138,35.3930130004882) [438](139,35.0568046569824) [439](140,34.5823554992675) [440](141,34.6861343383789) [441](142,34.395278930664) [442](143,34.5477142333984) [443](144,34.2278594970703) [444](145,34.2176094055175) [445](146,34.5024185180664) [446](147,33.4079780578613) [447](148,33.2507019042968) [448](149,33.3730163574218) [449](150,33.5519523620605) [450](151,33.6431579589843) [451](152,33.7373275756835) [452](153,32.8437881469726) [453](154,32.7951622009277) [454](155,32.9018745422363) [455](156,32.5688552856445) [456](157,32.5594635009765) [457](158,32.2602996826171) [458](159,32.3344650268554) [459](160,32.5526809692382) [460](161,31.9044475555419) [461](162,31.9629669189453) [462](163,31.7077407836914) [463](164,31.8282756805419) [464](165,31.3016395568847) [465](166,31.3248863220214) [466](167,31.4126300811767) [467](168,31.4238548278808) [468](169,31.2512950897216) [469](170,30.9933471679687) [470](171,31.1003913879394) [471](172,30.9836978912353) [472](173,30.7405281066894) [473](174,30.7918891906738) [474](175,30.5703086853027) [475](176,30.5192317962646) [476](177,30.4807052612304) [477](178,30.5145111083984) [478](179,30.4455986022949) [479](180,30.2384986877441) [480](181,30.2643527984619) [481](182,30.1937351226806) [482](183,30.1629314422607) [483](184,30.0649871826171) [484](185,30.0247344970703) [485](186,29.9694938659667) [486](187,29.9193840026855) [487](188,29.9507598876953) [488](189,29.9312667846679) [489](190,29.9032001495361) [490](191,29.8739509582519) [491](192,29.8693943023681) [492](193,29.8501548767089) [493](194,29.8199081420898) [494](195,29.814224243164) [495](196,29.8131637573242) [496](197,29.8030738830566) [497](198,29.7994728088378) [498](199,29.7989997863769) [499](200,29.7989997863769) [500]};

\addplot[color=clr6,mark='.',smooth, line width=0.25mm] coordinates {(0,123.922485351562) [1500](1,114.604362487792) [1501](2,108.796318054199) [1502](3,102.535385131835) [1503](4,98.8896026611328) [1504](5,94.8437042236328) [1505](6,91.1502838134765) [1506](7,88.5408020019531) [1507](8,85.2430572509765) [1508](9,85.5058670043945) [1509](10,80.9625549316406) [1510](11,77.7710418701171) [1511](12,78.9840698242187) [1512](13,78.2478485107421) [1513](14,74.4381866455078) [1514](15,72.385498046875) [1515](16,72.7592468261718) [1516](17,70.9909515380859) [1517](18,69.625503540039) [1518](19,68.6199874877929) [1519](20,66.3728637695312) [1520](21,65.1338806152343) [1521](22,66.2767333984375) [1522](23,63.9795989990234) [1523](24,64.6423568725586) [1524](25,63.0945968627929) [1525](26,63.9242515563964) [1526](27,60.0319862365722) [1527](28,61.61572265625) [1528](29,60.0317420959472) [1529](30,60.2892227172851) [1530](31,59.3270187377929) [1531](32,59.037727355957) [1532](33,57.7168807983398) [1533](34,57.1450691223144) [1534](35,56.1486129760742) [1535](36,56.4239425659179) [1536](37,56.9855804443359) [1537](38,55.4409866333007) [1538](39,54.356834411621) [1539](40,53.6743431091308) [1540](41,51.9019699096679) [1541](42,52.2901077270507) [1542](43,53.7125778198242) [1543](44,52.0759620666503) [1544](45,51.956485748291) [1545](46,50.7560539245605) [1546](47,50.3576507568359) [1547](48,50.8041610717773) [1548](49,48.6443099975585) [1549](50,49.3291130065918) [1550](51,49.58345413208) [1551](52,48.7009239196777) [1552](53,48.9540138244628) [1553](54,47.2989654541015) [1554](55,47.7972717285156) [1555](56,46.3048019409179) [1556](57,46.3836517333984) [1557](58,46.006965637207) [1558](59,45.6323471069335) [1559](60,44.7679901123046) [1560](61,43.8766403198242) [1561](62,45.20112991333) [1562](63,44.023208618164) [1563](64,43.1203079223632) [1564](65,43.5660095214843) [1565](66,43.1288681030273) [1566](67,42.8429679870605) [1567](68,42.9101943969726) [1568](69,43.07958984375) [1569](70,42.5853767395019) [1570](71,42.3579711914062) [1571](72,41.1955261230468) [1572](73,42.4234771728515) [1573](74,41.138687133789) [1574](75,41.060546875) [1575](76,41.31201171875) [1576](77,40.1100044250488) [1577](78,40.5817031860351) [1578](79,40.2138290405273) [1579](80,40.1642761230468) [1580](81,39.2781906127929) [1581](82,39.6233367919921) [1582](83,39.8049774169921) [1583](84,40.5334968566894) [1584](85,38.9463119506835) [1585](86,38.8045349121093) [1586](87,38.0946006774902) [1587](88,37.7783432006835) [1588](89,38.4944305419921) [1589](90,38.1818084716796) [1590](91,37.7006301879882) [1591](92,37.6993026733398) [1592](93,37.9392547607421) [1593](94,37.1392021179199) [1594](95,36.6073188781738) [1595](96,37.3518829345703) [1596](97,37.358528137207) [1597](98,36.6881065368652) [1598](99,36.9236679077148) [1599](100,35.989990234375) [1600](101,36.0794486999511) [1601](102,36.1139526367187) [1602](103,35.5958099365234) [1603](104,35.8699340820312) [1604](105,35.8830261230468) [1605](106,34.9605903625488) [1606](107,35.7620811462402) [1607](108,36.0648040771484) [1608](109,35.4828796386718) [1609](110,35.065933227539) [1610](111,34.8876228332519) [1611](112,34.5251808166503) [1612](113,34.8947525024414) [1613](114,34.0084533691406) [1614](115,33.89741897583) [1615](116,34.0628662109375) [1616](117,33.6708641052246) [1617](118,33.9323196411132) [1618](119,33.2077140808105) [1619](120,33.1413421630859) [1620](121,33.1737213134765) [1621](122,32.8002471923828) [1622](123,32.6568145751953) [1623](124,32.2964248657226) [1624](125,32.5565299987793) [1625](126,32.5690078735351) [1626](127,32.415657043457) [1627](128,32.5946197509765) [1628](129,32.5358276367187) [1629](130,32.5057525634765) [1630](131,32.2791481018066) [1631](132,31.4565391540527) [1632](133,31.8057746887207) [1633](134,31.7212467193603) [1634](135,31.319652557373) [1635](136,31.414566040039) [1636](137,30.9725608825683) [1637](138,30.9924583435058) [1638](139,30.5495910644531) [1639](140,30.7880668640136) [1640](141,30.6489601135253) [1641](142,30.0369625091552) [1642](143,30.4587421417236) [1643](144,30.1286315917968) [1644](145,30.0305595397949) [1645](146,29.8745765686035) [1646](147,30.1023139953613) [1647](148,29.7145462036132) [1648](149,29.531171798706) [1649](150,29.7444133758544) [1650](151,29.1522502899169) [1651](152,29.7189807891845) [1652](153,29.1255474090576) [1653](154,28.7760295867919) [1654](155,28.8208084106445) [1655](156,28.7210502624511) [1656](157,28.5575199127197) [1657](158,28.5884399414062) [1658](159,28.816650390625) [1659](160,28.5791587829589) [1660](161,28.4175205230712) [1661](162,28.3833827972412) [1662](163,28.0835075378417) [1663](164,28.2804012298584) [1664](165,28.1685771942138) [1665](166,27.9601554870605) [1666](167,28.0292205810546) [1667](168,27.9097175598144) [1668](169,27.9363536834716) [1669](170,27.7936973571777) [1670](171,27.6888275146484) [1671](172,27.5098533630371) [1672](173,27.590721130371) [1673](174,27.4741230010986) [1674](175,27.3850765228271) [1675](176,27.3648681640625) [1676](177,27.2984256744384) [1677](178,27.1995601654052) [1678](179,27.1071128845214) [1679](180,27.216739654541) [1680](181,27.0849075317382) [1681](182,27.099250793457) [1682](183,27.1029148101806) [1683](184,27.0210762023925) [1684](185,26.9701766967773) [1685](186,27.0033512115478) [1686](187,26.9234886169433) [1687](188,26.9431209564209) [1688](189,26.8473320007324) [1689](190,26.8741493225097) [1690](191,26.8894081115722) [1691](192,26.8391532897949) [1692](193,26.8712711334228) [1693](194,26.8338241577148) [1694](195,26.8233451843261) [1695](196,26.8200759887695) [1696](197,26.8191585540771) [1697](198,26.8297634124755) [1698](199,26.8354377746582) [1699](200,26.6925506591796) [1700]};

\addplot[color=clr3,mark='.',smooth, line width=0.25mm] coordinates {(0,137.919494628906) [600](1,136.781646728515) [601](2,122.624244689941) [602](3,126.093139648437) [603](4,116.030220031738) [604](5,125.613662719726) [605](6,107.561386108398) [606](7,105.981704711914) [607](8,103.873748779296) [608](9,103.848556518554) [609](10,103.434036254882) [610](11,95.2122039794921) [611](12,96.645637512207) [612](13,93.7975234985351) [613](14,103.379722595214) [614](15,98.9283599853515) [615](16,85.7344207763671) [616](17,83.1756286621093) [617](18,85.7342987060546) [618](19,86.2343139648437) [619](20,86.4203720092773) [620](21,86.9340362548828) [621](22,79.7194900512695) [622](23,75.6085662841796) [623](24,73.9759521484375) [624](25,79.4757690429687) [625](26,71.2929000854492) [626](27,75.8768768310546) [627](28,67.9258117675781) [628](29,69.9908447265625) [629](30,70.0374069213867) [630](31,62.9668884277343) [631](32,67.8189392089843) [632](33,66.0835800170898) [633](34,65.8376922607421) [634](35,61.5795059204101) [635](36,63.7616119384765) [636](37,58.5789489746093) [637](38,71.1673889160156) [638](39,57.9644241333007) [639](40,56.2294158935546) [640](41,59.4828338623046) [641](42,61.6752967834472) [642](43,57.8295097351074) [643](44,53.6527709960937) [644](45,61.6678161621093) [645](46,54.8096923828125) [646](47,50.5043106079101) [647](48,57.7568588256835) [648](49,50.769920349121) [649](50,53.3522415161132) [650](51,50.7598037719726) [651](52,47.3655319213867) [652](53,48.4193077087402) [653](54,49.3461227416992) [654](55,48.2271194458007) [655](56,48.2798156738281) [656](57,45.937759399414) [657](58,46.1490592956543) [658](59,48.8379898071289) [659](60,45.4217453002929) [660](61,42.635456085205) [661](62,44.1209754943847) [662](63,41.6265411376953) [663](64,42.4487380981445) [664](65,44.6922531127929) [665](66,43.811710357666) [666](67,46.2757186889648) [667](68,40.0037765502929) [668](69,39.6077117919921) [669](70,40.8754577636718) [670](71,40.1966476440429) [671](72,40.1546478271484) [672](73,40.6388931274414) [673](74,38.8037643432617) [674](75,38.1777915954589) [675](76,40.6416893005371) [676](77,38.7139739990234) [677](78,35.5326271057128) [678](79,36.6221084594726) [679](80,36.8915557861328) [680](81,34.9688453674316) [681](82,35.0368728637695) [682](83,35.1659774780273) [683](84,37.0985794067382) [684](85,34.1070098876953) [685](86,36.0811195373535) [686](87,34.248420715332) [687](88,35.4984397888183) [688](89,34.3716773986816) [689](90,35.2326583862304) [690](91,33.9782981872558) [691](92,34.6050872802734) [692](93,33.895767211914) [693](94,34.1443824768066) [694](95,34.8018798828125) [695](96,34.1129913330078) [696](97,33.2823181152343) [697](98,32.4355010986328) [698](99,32.8409347534179) [699](100,31.910005569458) [700](101,33.1797485351562) [701](102,32.3593215942382) [702](103,31.6893692016601) [703](104,32.581916809082) [704](105,31.6498222351074) [705](106,31.4846420288085) [706](107,31.2520713806152) [707](108,31.117919921875) [708](109,29.9942951202392) [709](110,31.4317321777343) [710](111,30.3888931274414) [711](112,29.8343601226806) [712](113,32.5995407104492) [713](114,29.8693447113037) [714](115,30.1960716247558) [715](116,29.541296005249) [716](117,30.9019546508789) [717](118,29.2122135162353) [718](119,28.9779357910156) [719](120,29.2183532714843) [720](121,30.2240676879882) [721](122,28.5520668029785) [722](123,28.7941780090332) [723](124,28.6866397857666) [724](125,28.2673454284667) [725](126,28.3789672851562) [726](127,28.7930107116699) [727](128,28.5150260925292) [728](129,27.3858909606933) [729](130,27.7700843811035) [730](131,27.4459838867187) [731](132,27.8168869018554) [732](133,27.0580158233642) [733](134,26.8809452056884) [734](135,27.0919227600097) [735](136,26.9539222717285) [736](137,26.9532661437988) [737](138,26.7882423400878) [738](139,26.4222946166992) [739](140,25.8480911254882) [740](141,26.053108215332) [741](142,25.6417083740234) [742](143,25.9527797698974) [743](144,25.7374248504638) [744](145,25.5284690856933) [745](146,25.1045150756835) [746](147,25.5898094177246) [747](148,25.3310070037841) [748](149,25.4073028564453) [749](150,25.3672676086425) [750](151,24.8116416931152) [751](152,24.750244140625) [752](153,24.462287902832) [753](154,24.3219261169433) [754](155,26.2133331298828) [755](156,24.3362731933593) [756](157,24.3512058258056) [757](158,24.2561721801757) [758](159,24.3776149749755) [759](160,23.8340644836425) [760](161,23.8512134552001) [761](162,23.72509765625) [762](163,23.731761932373) [763](164,23.5006694793701) [764](165,23.5484828948974) [765](166,23.3843631744384) [766](167,23.4677581787109) [767](168,23.0534152984619) [768](169,23.2703838348388) [769](170,23.0292510986328) [770](171,23.078758239746) [771](172,22.9958763122558) [772](173,22.8945522308349) [773](174,22.6556682586669) [774](175,22.8111305236816) [775](176,22.7097377777099) [776](177,22.6789684295654) [777](178,22.6418800354003) [778](179,22.5381202697753) [779](180,22.5255489349365) [780](181,22.3505859375) [781](182,22.3287277221679) [782](183,22.3515243530273) [783](184,22.2749900817871) [784](185,22.2925376892089) [785](186,22.2377338409423) [786](187,22.1880855560302) [787](188,22.1887168884277) [788](189,22.1584682464599) [789](190,22.1388721466064) [790](191,22.1125679016113) [791](192,22.1003913879394) [792](193,22.0865573883056) [793](194,22.0827674865722) [794](195,22.0737190246582) [795](196,22.0629253387451) [796](197,22.0591621398925) [797](198,22.0571403503417) [798](199,22.0570774078369) [799](200,22.0570774078369) [800]};

\addplot[color=clr7,mark='.',smooth, line width=0.25mm] coordinates {(0,118.880950927734) [1800](1,109.419075012207) [1801](2,102.430068969726) [1802](3,95.6804046630859) [1803](4,93.1300201416015) [1804](5,90.236328125) [1805](6,87.182632446289) [1806](7,85.5429992675781) [1807](8,83.9642486572265) [1808](9,82.1343002319336) [1809](10,81.3818817138671) [1810](11,79.4234619140625) [1811](12,77.8119354248046) [1812](13,77.6605224609375) [1813](14,76.7805252075195) [1814](15,76.0213165283203) [1815](16,74.3598556518554) [1816](17,72.8290405273437) [1817](18,71.4181976318359) [1818](19,71.1163253784179) [1819](20,69.8159790039062) [1820](21,68.8293991088867) [1821](22,67.707649230957) [1822](23,66.3991241455078) [1823](24,67.6613311767578) [1824](25,66.4583587646484) [1825](26,65.8975372314453) [1826](27,64.2528228759765) [1827](28,62.8311042785644) [1828](29,62.6640129089355) [1829](30,63.1481704711914) [1830](31,62.9822273254394) [1831](32,60.8149795532226) [1832](33,59.8552627563476) [1833](34,58.8761978149414) [1834](35,60.3736877441406) [1835](36,57.4321823120117) [1836](37,58.6708374023437) [1837](38,57.1501502990722) [1838](39,57.362319946289) [1839](40,57.6125946044921) [1840](41,55.5374488830566) [1841](42,54.9725914001464) [1842](43,55.0168418884277) [1843](44,54.6508178710937) [1844](45,55.6385040283203) [1845](46,54.0798645019531) [1846](47,53.2624168395996) [1847](48,53.0779457092285) [1848](49,53.2701339721679) [1849](50,54.2857437133789) [1850](51,51.9531097412109) [1851](52,51.8045578002929) [1852](53,50.4576339721679) [1853](54,51.2580413818359) [1854](55,50.7367095947265) [1855](56,50.0237503051757) [1856](57,50.412368774414) [1857](58,49.9665985107421) [1858](59,50.1788520812988) [1859](60,49.5186462402343) [1860](61,48.9761276245117) [1861](62,49.287254333496) [1862](63,49.3540954589843) [1863](64,47.811725616455) [1864](65,47.5664176940918) [1865](66,49.6317405700683) [1866](67,47.2432250976562) [1867](68,47.9621353149414) [1868](69,46.4932250976562) [1869](70,47.5620536804199) [1870](71,46.6885070800781) [1871](72,46.4017257690429) [1872](73,45.914207458496) [1873](74,45.2577896118164) [1874](75,44.7482528686523) [1875](76,45.9745025634765) [1876](77,45.1080474853515) [1877](78,46.793399810791) [1878](79,44.9582901000976) [1879](80,45.4491233825683) [1880](81,43.9784049987793) [1881](82,43.733528137207) [1882](83,43.3433609008789) [1883](84,43.5238761901855) [1884](85,43.688621520996) [1885](86,43.0173645019531) [1886](87,42.7813720703125) [1887](88,43.5156440734863) [1888](89,42.3579940795898) [1889](90,42.1832962036132) [1890](91,42.935375213623) [1891](92,42.0748786926269) [1892](93,42.6253585815429) [1893](94,41.5052032470703) [1894](95,41.9570274353027) [1895](96,41.0527610778808) [1896](97,40.574291229248) [1897](98,40.5867538452148) [1898](99,40.2427291870117) [1899](100,40.2451248168945) [1900](101,41.2039794921875) [1901](102,40.1259002685546) [1902](103,39.7150497436523) [1903](104,39.6839294433593) [1904](105,39.5749702453613) [1905](106,39.7719039916992) [1906](107,39.7141189575195) [1907](108,39.8083114624023) [1908](109,38.7072601318359) [1909](110,38.8124237060546) [1910](111,39.3839721679687) [1911](112,39.1246910095214) [1912](113,38.4921722412109) [1913](114,39.4962158203125) [1914](115,38.8528518676757) [1915](116,37.9232597351074) [1916](117,38.329734802246) [1917](118,38.788818359375) [1918](119,39.096923828125) [1919](120,37.6171188354492) [1920](121,38.1040077209472) [1921](122,37.5537948608398) [1922](123,37.3188934326171) [1923](124,37.6822204589843) [1924](125,37.6034431457519) [1925](126,37.1971740722656) [1926](127,36.7176551818847) [1927](128,36.876968383789) [1928](129,37.3233947753906) [1929](130,36.8823318481445) [1930](131,36.4055023193359) [1931](132,36.698875427246) [1932](133,38.517234802246) [1933](134,36.3828659057617) [1934](135,36.1124305725097) [1935](136,35.6054534912109) [1936](137,36.4662132263183) [1937](138,35.3305282592773) [1938](139,34.8597717285156) [1939](140,35.3456878662109) [1940](141,35.4929428100585) [1941](142,35.0660133361816) [1942](143,34.9633522033691) [1943](144,35.3171157836914) [1944](145,34.5454025268554) [1945](146,34.5578460693359) [1946](147,34.8169021606445) [1947](148,34.3572082519531) [1948](149,34.4025993347168) [1949](150,34.2001991271972) [1950](151,33.823616027832) [1951](152,33.9872131347656) [1952](153,34.3284835815429) [1953](154,34.2276382446289) [1954](155,33.5849800109863) [1955](156,34.0847396850585) [1956](157,33.7123260498046) [1957](158,33.874828338623) [1958](159,33.9892807006835) [1959](160,33.9095764160156) [1960](161,33.7398719787597) [1961](162,33.6248092651367) [1962](163,34.0315361022949) [1963](164,33.7963142395019) [1964](165,33.088150024414) [1965](166,33.6649589538574) [1966](167,33.0654296875) [1967](168,32.936538696289) [1968](169,33.2223320007324) [1969](170,33.33544921875) [1970](171,32.8149185180664) [1971](172,32.6826782226562) [1972](173,32.7306976318359) [1973](174,32.8914794921875) [1974](175,32.7689361572265) [1975](176,32.4997100830078) [1976](177,32.7989082336425) [1977](178,32.6644439697265) [1978](179,32.736099243164) [1979](180,32.8928871154785) [1980](181,32.5216522216796) [1981](182,33.1823539733886) [1982](183,33.5497512817382) [1983](184,33.3430633544921) [1984](185,33.68599319458) [1985](186,32.2518920898437) [1986](187,32.9165954589843) [1987](188,32.9370956420898) [1988](189,32.1938018798828) [1989](190,33.1129302978515) [1990](191,33.1444816589355) [1991](192,32.763500213623) [1992](193,32.0753555297851) [1993](194,32.1225433349609) [1994](195,32.5567321777343) [1995](196,32.0259628295898) [1996](197,32.0570068359375) [1997](198,33.5404739379882) [1998](199,32.0822296142578) [1999](200,32.0861663818359) [2000]};

\addplot[color=clr4,mark='.',smooth, line width=0.25mm] coordinates {(0,135.662414550781) [900](1,126.413192749023) [901](2,123.043090820312) [902](3,122.361206054687) [903](4,119.19856262207) [904](5,113.4560546875) [905](6,112.866752624511) [906](7,109.274169921875) [907](8,109.501098632812) [908](9,103.881721496582) [909](10,99.9273529052734) [910](11,93.7986145019531) [911](12,93.6110382080078) [912](13,90.5462646484375) [913](14,90.6158905029296) [914](15,86.639663696289) [915](16,83.5966339111328) [916](17,82.2625122070312) [917](18,78.0940933227539) [918](19,76.4490432739257) [919](20,71.6122970581054) [920](21,77.0772323608398) [921](22,72.2347259521484) [922](23,71.3508148193359) [923](24,74.1249084472656) [924](25,70.5917663574218) [925](26,67.7499084472656) [926](27,63.0406799316406) [927](28,69.4322662353515) [928](29,70.8783721923828) [929](30,67.0414047241211) [930](31,68.824951171875) [931](32,75.4076232910156) [932](33,63.8053894042968) [933](34,59.1217346191406) [934](35,63.4605026245117) [935](36,61.6684188842773) [936](37,58.5588989257812) [937](38,63.1000137329101) [938](39,58.9171142578125) [939](40,55.5290336608886) [940](41,53.4246444702148) [941](42,53.9598083496093) [942](43,54.4307060241699) [943](44,53.3797454833984) [944](45,53.9596328735351) [945](46,54.8967781066894) [946](47,51.2560348510742) [947](48,53.7059478759765) [948](49,51.7679748535156) [949](50,48.8220977783203) [950](51,50.366828918457) [951](52,52.7525672912597) [952](53,50.2445335388183) [953](54,47.7405624389648) [954](55,47.5227508544921) [955](56,47.5073852539062) [956](57,46.2077407836914) [957](58,46.2140045166015) [958](59,45.7079811096191) [959](60,45.4468841552734) [960](61,44.798095703125) [961](62,47.1315231323242) [962](63,46.2756805419921) [963](64,47.1983413696289) [964](65,45.121337890625) [965](66,42.7749862670898) [966](67,44.2000160217285) [967](68,45.2712593078613) [968](69,43.7633895874023) [969](70,42.1181335449218) [970](71,43.3629837036132) [971](72,41.5319862365722) [972](73,42.7147064208984) [973](74,41.2357177734375) [974](75,41.1114120483398) [975](76,42.5043792724609) [976](77,42.6259231567382) [977](78,41.2059326171875) [978](79,40.518138885498) [979](80,41.0764007568359) [980](81,40.6126708984375) [981](82,39.077537536621) [982](83,39.6828536987304) [983](84,40.3609161376953) [984](85,38.8106842041015) [985](86,39.468147277832) [986](87,41.5198287963867) [987](88,38.8090744018554) [988](89,39.121597290039) [989](90,37.2087631225585) [990](91,38.731689453125) [991](92,38.0718688964843) [992](93,36.2557067871093) [993](94,38.9756164550781) [994](95,37.4807739257812) [995](96,38.9572525024414) [996](97,35.919319152832) [997](98,36.7830238342285) [998](99,35.7106323242187) [999](100,36.4777069091796) [1000](101,34.7016639709472) [1001](102,35.265884399414) [1002](103,34.6848640441894) [1003](104,35.1517028808593) [1004](105,35.3134994506835) [1005](106,34.0087203979492) [1006](107,35.7431259155273) [1007](108,34.2627334594726) [1008](109,35.2171020507812) [1009](110,34.8921089172363) [1010](111,34.0078125) [1011](112,34.5190277099609) [1012](113,33.1073837280273) [1013](114,32.4580116271972) [1014](115,33.4666061401367) [1015](116,33.1434707641601) [1016](117,32.5744400024414) [1017](118,32.8024291992187) [1018](119,32.5178527832031) [1019](120,33.2988815307617) [1020](121,32.0420341491699) [1021](122,31.6913108825683) [1022](123,31.2803916931152) [1023](124,32.12744140625) [1024](125,32.1027450561523) [1025](126,32.0222969055175) [1026](127,31.1791400909423) [1027](128,30.4422454833984) [1028](129,30.9481544494628) [1029](130,30.4777507781982) [1030](131,30.107421875) [1031](132,30.21484375) [1032](133,29.9677276611328) [1033](134,29.5672874450683) [1034](135,29.2446327209472) [1035](136,29.4895076751709) [1036](137,29.2463035583496) [1037](138,29.194164276123) [1038](139,29.0412998199462) [1039](140,29.2768287658691) [1040](141,28.3218994140625) [1041](142,28.9470748901367) [1042](143,28.4531421661376) [1043](144,28.7586116790771) [1044](145,28.1519012451171) [1045](146,28.0480041503906) [1046](147,27.9738006591796) [1047](148,27.9572830200195) [1048](149,27.8426055908203) [1049](150,27.2681941986084) [1050](151,27.317398071289) [1051](152,27.1265354156494) [1052](153,27.052225112915) [1053](154,26.7404403686523) [1054](155,26.7863960266113) [1055](156,27.0355548858642) [1056](157,26.8762512207031) [1057](158,26.4318447113037) [1058](159,26.9231300354003) [1059](160,26.3211402893066) [1060](161,26.1556453704834) [1061](162,25.925588607788) [1062](163,25.8900032043457) [1063](164,26.1336574554443) [1064](165,25.772102355957) [1065](166,25.6628608703613) [1066](167,25.7077941894531) [1067](168,25.4758377075195) [1068](169,25.331895828247) [1069](170,25.3836441040039) [1070](171,25.3714065551757) [1071](172,25.1479415893554) [1072](173,25.0122528076171) [1073](174,24.9813404083251) [1074](175,24.9641456604003) [1075](176,24.9355487823486) [1076](177,24.8928146362304) [1077](178,24.7711086273193) [1078](179,24.7366619110107) [1079](180,24.5942993164062) [1080](181,24.4948215484619) [1081](182,24.5446434020996) [1082](183,24.5286388397216) [1083](184,24.387523651123) [1084](185,24.3350982666015) [1085](186,24.3613166809082) [1086](187,24.31148147583) [1087](188,24.2700805664062) [1088](189,24.2268638610839) [1089](190,24.2166576385498) [1090](191,24.2068061828613) [1091](192,24.190408706665) [1092](193,24.1753768920898) [1093](194,24.150089263916) [1094](195,24.1433486938476) [1095](196,24.1359405517578) [1096](197,24.1344318389892) [1097](198,24.133804321289) [1098](199,24.1336135864257) [1099](200,24.1336135864257) [1100]};

\addplot[color=clr8,mark='.',smooth , line width=0.25mm] coordinates {(0,119.90364074707) [2100](1,101.569869995117) [2101](2,93.5553512573242) [2102](3,85.8389892578125) [2103](4,83.1685333251953) [2104](5,77.7892913818359) [2105](6,76.1154327392578) [2106](7,72.7016143798828) [2107](8,73.3598175048828) [2108](9,71.8559646606445) [2109](10,68.6951599121093) [2110](11,66.222541809082) [2111](12,66.0070190429687) [2112](13,64.3246994018554) [2113](14,63.2469177246093) [2114](15,62.3570022583007) [2115](16,61.2963905334472) [2116](17,61.3122787475585) [2117](18,59.7909431457519) [2118](19,60.140037536621) [2119](20,57.6462745666503) [2120](21,58.5732841491699) [2121](22,58.4506340026855) [2122](23,56.2618103027343) [2123](24,56.6962127685546) [2124](25,56.3198013305664) [2125](26,55.6997604370117) [2126](27,53.3223190307617) [2127](28,53.0489768981933) [2128](29,53.3578491210937) [2129](30,54.4181594848632) [2130](31,50.9724960327148) [2131](32,52.2833023071289) [2132](33,52.007583618164) [2133](34,50.9599838256835) [2134](35,50.8333282470703) [2135](36,50.0531387329101) [2136](37,50.6862983703613) [2137](38,50.4715995788574) [2138](39,49.0557174682617) [2139](40,49.6941299438476) [2140](41,48.4183502197265) [2141](42,47.3905029296875) [2142](43,46.9597625732421) [2143](44,49.3606872558593) [2144](45,48.0415878295898) [2145](46,47.702522277832) [2146](47,46.9963607788085) [2147](48,46.7969551086425) [2148](49,45.828140258789) [2149](50,46.7169342041015) [2150](51,46.6414451599121) [2151](52,47.3693084716796) [2152](53,44.9845962524414) [2153](54,43.7780456542968) [2154](55,43.6995468139648) [2155](56,43.5531616210937) [2156](57,43.4462699890136) [2157](58,42.8849182128906) [2158](59,42.1686172485351) [2159](60,42.386173248291) [2160](61,42.5156173706054) [2161](62,43.451602935791) [2162](63,42.8015670776367) [2163](64,41.7409706115722) [2164](65,41.448486328125) [2165](66,41.9313621520996) [2166](67,41.9647560119628) [2167](68,41.8399314880371) [2168](69,41.258071899414) [2169](70,40.0512580871582) [2170](71,39.5424079895019) [2171](72,39.3699531555175) [2172](73,39.2028999328613) [2173](74,39.3412857055664) [2174](75,39.1720428466796) [2175](76,39.1408424377441) [2176](77,38.3683624267578) [2177](78,38.5960388183593) [2178](79,39.4494552612304) [2179](80,38.0313491821289) [2180](81,37.3705291748046) [2181](82,37.8934516906738) [2182](83,37.3754806518554) [2183](84,36.8231506347656) [2184](85,36.6310348510742) [2185](86,36.2956848144531) [2186](87,36.2967300415039) [2187](88,38.1699180603027) [2188](89,36.0654602050781) [2189](90,36.3019638061523) [2190](91,36.3008270263671) [2191](92,35.800178527832) [2192](93,36.3919448852539) [2193](94,35.2941665649414) [2194](95,35.2388954162597) [2195](96,34.9944229125976) [2196](97,35.9532470703125) [2197](98,34.6561813354492) [2198](99,34.6003646850585) [2199](100,35.0146179199218) [2200](101,34.8231658935546) [2201](102,34.7543182373046) [2202](103,34.5875167846679) [2203](104,34.1142654418945) [2204](105,34.4106063842773) [2205](106,33.8633728027343) [2206](107,33.5771484375) [2207](108,32.7642097473144) [2208](109,32.6980895996093) [2209](110,32.6013259887695) [2210](111,32.7679252624511) [2211](112,32.7627868652343) [2212](113,32.8091812133789) [2213](114,33.4331512451171) [2214](115,32.0800094604492) [2215](116,32.7489166259765) [2216](117,31.7555332183837) [2217](118,31.615737915039) [2218](119,31.4896202087402) [2219](120,31.4297504425048) [2220](121,31.8826255798339) [2221](122,31.1260719299316) [2222](123,31.1536064147949) [2223](124,30.8992233276367) [2224](125,31.0291633605957) [2225](126,30.7445487976074) [2226](127,30.0078544616699) [2227](128,30.476505279541) [2228](129,30.4878349304199) [2229](130,30.1368141174316) [2230](131,29.8221893310546) [2231](132,29.3402595520019) [2232](133,29.4984416961669) [2233](134,29.3355255126953) [2234](135,29.4883022308349) [2235](136,29.0108757019042) [2236](137,29.0055465698242) [2237](138,29.239782333374) [2238](139,28.9083003997802) [2239](140,28.8792457580566) [2240](141,28.4775199890136) [2241](142,28.1909465789794) [2242](143,28.3820648193359) [2243](144,28.3885078430175) [2244](145,27.9658966064453) [2245](146,27.9872550964355) [2246](147,27.7857818603515) [2247](148,27.6818199157714) [2248](149,27.6900291442871) [2249](150,27.6950645446777) [2250](151,27.5950603485107) [2251](152,27.2089958190917) [2252](153,27.47603225708) [2253](154,27.120002746582) [2254](155,27.0468711853027) [2255](156,27.0445480346679) [2256](157,26.6570167541503) [2257](158,26.6537933349609) [2258](159,26.5203742980957) [2259](160,26.4785919189453) [2260](161,26.2926216125488) [2261](162,26.2059211730957) [2262](163,26.3635101318359) [2263](164,26.1582565307617) [2264](165,25.8848571777343) [2265](166,25.8799476623535) [2266](167,25.813585281372) [2267](168,25.7373180389404) [2268](169,25.6214561462402) [2269](170,25.6047134399414) [2270](171,25.5009155273437) [2271](172,25.5149326324462) [2272](173,25.3242568969726) [2273](174,25.2951412200927) [2274](175,25.1927318572998) [2275](176,25.1601123809814) [2276](177,25.0589179992675) [2277](178,25.0304298400878) [2278](179,25.0012474060058) [2279](180,25.0296211242675) [2280](181,24.8848857879638) [2281](182,24.887128829956) [2282](183,24.797061920166) [2283](184,24.8365631103515) [2284](185,24.7625560760498) [2285](186,24.7478446960449) [2286](187,24.7450828552246) [2287](188,24.7081069946289) [2288](189,24.6976871490478) [2289](190,24.6816482543945) [2290](191,24.6496067047119) [2291](192,24.6470336914062) [2292](193,24.6286392211914) [2293](194,24.6236228942871) [2294](195,24.6147918701171) [2295](196,24.6211585998535) [2296](197,24.6203708648681) [2297](198,24.6028671264648) [2298](199,24.6092834472656) [2299](200,24.6028671264648) [2300]};

\legend{STLight, SimVP+IncepU, ConvLSTM, SimVP+gSTA-S, PredRNN++, SimVP+ConvMixer, PredRNN.V2, TAU}
\end{axis}
\end{tikzpicture}

%% file: plots/dimvsdepth.tex
\begin{tikzpicture}

\definecolor{clr1}{RGB}{255,0,0}
\definecolor{clr2}{RGB}{0,255,0}
\definecolor{clr3}{RGB}{255,0,255}
\definecolor{clr4}{RGB}{0,0,255}
\definecolor{clr5}{RGB}{255,127,127}

\begin{axis}[
    xlabel={Parameters (in millions)},
    ylabel={MSE},
    xmin=10, xmax=70,
    ymin=21, ymax=25.9,
    legend pos=north east,
    legend style={font=\large},
    ymajorgrids=true,
    legend columns=2,
    grid style=dashed
]

\addplot[color=clr1,mark=*,  line width=0.25mm]
    coordinates {
(10.856,25.6957) [0]
(13.018,24.6801) [1]
(15.18,23.8692) [2]
(17.342,23.3292) [3]
(19.504,23.1634) [4]
(21.666,22.9401) [5]
(23.828,22.7314) [6]
(25.99,22.6384) [7]
(28.152,22.6172) [8]
(30.314,22.6544) [9]
};

\addplot[color=clr2,mark=*, line width=0.25mm]
    coordinates {
(15.427,24.5663) [10]
(18.501,23.6514) [11]
(21.575,22.9628) [12]
(24.65,22.2313) [13]
(27.724,22.007) [14]
(30.799,21.8168) [15]
(33.873,21.7265) [16]
(36.947,21.6276) [17]
(40.022,21.6323) [18]
};

\addplot[color=clr3,mark=*, line width=0.25mm]
    coordinates {
(20.798,24.1499) [19]
(24.945,23.1457) [20]
(29.091,22.4202) [21]
(33.238,21.8308) [22]
(37.385,21.6834) [23]
(41.532,21.5308) [24]
(45.679,21.4246) [25]
(49.825,21.3976) [26]
};

\addplot[color=clr4,mark=*, line width=0.25mm]
    coordinates {
(32.348,22.8462) [27]
(37.727,22.1723) [28]
(43.106,21.6184) [29]
(48.486,21.3122) [30]
(53.865,21.2794) [31]
(59.244,21.252) [32]
(64.623,21.1368) [33]
};

\addplot[color=clr5,mark=*, line width=0.25mm]
    coordinates {
(40.712,22.5699) [34]
(47.483,21.8942) [35]
(54.255,21.3303) [36]
(61.026,21.1909) [37]
(67.798,21.0842) [38]
};

\legend{$d$=1000, $d$=1200, $d$=1400, $d$=1600, $d$=1800}
\end{axis}
\end{tikzpicture}

%% file: plots/overlapping.tex
\begin{tikzpicture}

\begin{axis}[
    xlabel={Parameters (in millions)},
    ylabel={MSE},
    xmin=10, xmax=70,
    ymin=20, ymax=27,
    legend pos=north east,
    legend style={font=\large},
    ymajorgrids=true,
    grid style=dashed,
]
\addplot[scatter,only marks,scatter src=explicit symbolic,mark=*,draw=blue]
    coordinates {
(12.771,25.67) [1]
(14.18,25.5) [2]
(15.59,25.3) [3]
(17,25.23) [4]
(18.409,25.16) [5]
(18.573,24.42) [6]
(19.564,24.13) [7]
(20.882,24.33) [8]
(21.726,23.94) [9]
(23.888,23.82) [10]
(25.029,23.75) [11]
(26.05,23.56) [12]
(27.796,23.7) [13]
(28.212,23.72) [14]
(30.374,23.02) [15]
(30.871,22.92) [16]
(33.945,22.69) [17]
(37.019,22.58) [18]
(37.469,22.97) [19]
(40.094,22.27) [20]
(41.616,22.09) [21]
(45.763,21.91) [22]
(48.582,22.21) [23]
(49.909,21.82) [24]
(53.961,21.99) [25]
(59.34,21.65) [26]
(61.134,21.56) [27]
(64.719,21.38) [28]
(67.906,21.03) [29] 
};

\addplot[only marks,scatter src=explicit symbolic,mark=*,draw=red,mark options={color=red},]
    coordinates {
(12.771,26.85) [31]
(14.18,26.44) [32]
(15.59,26.14) [33]
(17,26.11) [34]
(18.409,25.42) [35]
(18.573,25.23) [36]
(19.564,24.53) [37]
(20.882,25.13) [38]
(21.726,24.44) [39]
(23.888,24.14) [40]
(25.029,23.95) [41]
(26.05,24.19) [42]
(27.796,23.17) [43]
(28.212,23.54) [44]
(30.374,23.53) [45]
(30.871,22.77) [46]
(33.945,22.57) [47]
(37.019,22.4) [48]
(37.469,22.07) [49]
(40.094,22.29) [50]
(41.616,21.49) [51]
(45.763,21.39) [52]
(48.582,21) [53]
(49.909,21.19) [54]
(53.961,20.43) [55]
(59.34,20.56) [56]
(61.134,20.1) [57]
(64.719,20.52) [58]
(67.906,20.04) [59]
};

\legend{Without overlapping: $O=0$, With overlapping: $O=2$}
\end{axis}
\end{tikzpicture}

%% file: sec/5_conclusions.tex
\section{Conclusions}
 We introduce STLight, an innovative STL architecture that either matches or surpasses the accuracy of the state-of-the-art recurrent methods, while also offering greater efficiency than recurrent-free methods. This work challenges the prevailing perspectives in STL, showing that convolutions alone can effectively joint capture spatial and temporal dependencies, eliminating the need for complex additional modules and strategies. We achieve these results by introducing spatio-temporal patches, an enhanced representation of a sequence of frames, that joint integrates both temporal and spatial information, and by surpassing traditional \textit{Spatial-Temporal-Spatial} paradigm towards a more comprehensive framework where both spatial and temporal information are jointly integrated. Furthermore, the reduced computational demand facilitates scaling to a wider range of scenarios. We believe this contribution could serve as a robust baseline and inspire future research in the field.

%% file: sec/x_appendix.tex
\clearpage
\setcounter{page}{1}
\setcounter{section}{0}
\renewcommand*{\thesection}{\Alph{section}}

\lstdefinestyle{mystyle}{
    commentstyle=\color{codegreen},
    keywordstyle=\color{cvprblue},
    numberstyle=\tiny\color{codegray},
    stringstyle=\color{codepurple},
    basicstyle=\ttfamily\small,
    breakatwhitespace=false,         
    breaklines=true,                 
    captionpos=b,                    
    keepspaces=true,                 
    numbers=left,                    
    numbersep=1pt,                  
    showspaces=false,                
    showstringspaces=false,
    showtabs=false,                  
    tabsize=1
}
\definecolor{cvprblue}{rgb}{0.21,0.49,0.74}
\definecolor{codegreen}{rgb}{0,0.6,0}
\definecolor{codegray}{rgb}{0.5,0.5,0.5}
\definecolor{codepurple}{rgb}{0.58,0,0.82}
\definecolor{backcolour}{rgb}{0.95,0.95,0.92}
\lstset{style=mystyle}

\pgfplotstableset{
    /color cells/min/.initial=0,
    /color cells/max/.initial=1000,
    /color cells/textcolor/.initial=,
    %
    color cells/.code={%
        \pgfqkeys{/color cells}{#1}%
        \pgfkeysalso{%
            postproc cell content/.code={%
                \begingroup
                %
                \pgfkeysgetvalue{/pgfplots/table/@preprocessed cell content}\value
\ifx\value\empty
\endgroup
\else
                \pgfmathfloatparsenumber{\value}%
                \pgfmathfloattofixed{\pgfmathresult}%
                \let\value=\pgfmathresult
                %
                \pgfplotscolormapaccess
                    [\pgfkeysvalueof{/color cells/min}:\pgfkeysvalueof{/color cells/max}]%
                    {\value}%
                    {\pgfkeysvalueof{/pgfplots/colormap name}}%
                %
                \pgfkeysgetvalue{/pgfplots/table/@cell content}\typesetvalue
                \pgfkeysgetvalue{/color cells/textcolor}\textcolorvalue
                %
                \toks0=\expandafter{\typesetvalue}%
                \xdef\temp{%
                    \noexpand\pgfkeysalso{%
                        @cell content={%
                            \noexpand\cellcolor[rgb]{\pgfmathresult}%
                            \noexpand\definecolor{mapped color}{rgb}{\pgfmathresult}%
                            \ifx\textcolorvalue\empty
                            \else
                                \noexpand\color{\textcolorvalue}%
                            \fi
                            \the\toks0 %
                        }%
                    }%
                }%
                \endgroup
                \temp
\fi
            }%
        }%
    }
}

The supplementary material provides a comprehensive analysis of the STLight method. Each section contributes unique insights:

In Section \ref{adictional_exp}, we extend our evaluation by comparing our model with results from others published STL papers, against longer training duration, and presenting an ablation study on the impact of varying training hyperparameters.

In Section \ref{sec:complexity}, we analyze STLight's complexity, breaking down its components and parameter counts.

Section \ref{sec:kernel_size} explores optimal kernel sizes for STLight's convolutional stages.

In Section \ref{sec:weight_init}, we investigate the impact of weight initialization schemes on training stability.

Section \ref{sec:optimal_settings} focuses on optimal training settings.

In Section \ref{sec:decoding_operations}, we compare decoding operations' impact on model stability and performance.

Finally, Section \ref{sec:STLight_code} presents the full STLight implementation.

\section{Additional Evaluation Results}\label{adictional_exp}
In Section 4, we used the OpenSTL benchmark to compare our results with public and reproducible outcomes on established benchmark datasets. However, public libraries like OpenSTL do not fully guarantee (1) the correctness of the implementations, (2) the adherence to the original training protocols of each baseline, or (3) the optimality of the default standard training parameters used for learning. 

Thus, in this section, we address these limitations. 

In Table \ref{addictional_table}, we compare our model with results published in the literature for relevant STL models on the MMNIST, TaxiBJ, and KTH datasets. While all baseline results in Table 3 are obtained under uniform training settings and protocols, Table \ref{addictional_table} lacks this standardization. Given that each model in Table \ref{addictional_table} is trained for different (and not always reported) durations, we trained our STLight baseline using the same hyperparameters as in Table 2, but with extended training durations: 2000 epochs for MMNIST and 150 epochs for KTH. Our model still outperforms the other baselines, confirming the OpenSTL benchmark results from Table 3.

\begin{table}[h!] 
\centering
\caption{Comparison of our model results and the results published for each model in literature Across MMNIST, TaxiBJ and KTH datasets. Our model have been trained using hyperparameters of Table 2, except for the training duration that have been extended to 2000 epochs for MMNIST and 150 epochs for KTH. }
\resizebox{0.5\textwidth}{!}{%
\begin{tabular}{@{}lccc@{}}
\toprule
\textbf{Model} & \multicolumn{1}{c}{\textbf{MMNIST (MSE $\downarrow$)}} & \multicolumn{1}{c}{\textbf{TaxiBJ (MSE $\downarrow$)}} & \multicolumn{1}{c}{\textbf{KTH (SSIM $\uparrow$)}} \\
 &&& \textbf{10 $\rightarrow$ 20}  \\ \midrule
ConvLSTM \cite{shi2015convolutional} & 103.3 & 48.5 & 0.712  \\
VPTR-NAR \cite{ye2022vptr} & 107.2 & - & 0.859  \\
VPTR-FAR \cite{ye2022vptr} & 63.6 & - & 0.879  \\
PredRNN \cite{wang2017predrnn} & 56.8 & 46.4 & 0.839  \\
PredRNN++ \cite{wang2018predrnn++} & 46.5 & 44.8 & 0.865  \\
MIM \cite{wang2019memory} & 44.2 & 42.9 & -  \\
E3D-LSTM \cite{wang2018eidetic} & 41.3 & 43.2 & 0.879  \\
MAU \cite{chang2021mau} & 29.5 & - & -  \\
PhyDNet \cite{guen2020disentangling} & 24.4 & 41.9 & -  \\
Crevnet \cite{yu2019crevnet} & 22.3 & - & -  \\
PredRNNv2 \cite{wang2022predrnn} & 48.8 & - & - \\
IAM4VP \cite{seo2023implicit} & 15.3 & 37.2 & -  \\
TAU \cite{tan2023temporal} & 19.8 & 34.4 & 0.911  \\
\midrule
STLight-L (Ours) & \textbf{14.72} &\textbf{ 30.87}& \textbf{0.9113 }    \\   
\bottomrule
\end{tabular}%
} \label{addictional_table}
\end{table}

In Table \ref{long_runs}, we extend our model evaluation to the 2000-epoch OpenSTL MMNIST results. Notably, our model corroborates the findings discussed in Section 4.2.1, demonstrating superior performance and efficiency compared to the OpenSTL baselines across both accuracy and computational metrics. Specifically, even with extended training time, STLight-L achieves the best trade-off between accuracy and computational cost, with a significantly lower MSE (14.77) and MAE (47.17), while maintaining a high SSIM (0.9686). As discussed in Section 4.2.1, our model continues to outperform recurrent architectures like PredRNN++ and MIM in accuracy while requiring fewer FLOPs, and recurrent-free architectures like TAU and SimVP with fewer parameters, establishing STLight-L as both a highly performant and resource-efficient solution.

\begin{table}[h]
\centering
\caption{Quantitative results comparing our model (STLight-L) against OpenSTL baselines, trained for 2000 epochs using the training settings from Table 2. The table reports both accuracy metrics (MSE, MAE, SSIM) and computational metrics (number of parameters, FLOPs) under equivalent training and evaluation conditions on the MMNIST dataset}
 \resizebox{0.5\textwidth}{!}{
\begin{tabular}{lccccc}
\toprule
Model & \# Params & FLOPs & MSE & MAE & SSIM \\
\midrule
ConvLSTM-S & 15.0M & 56.8G & 22.41 & 73.07 & 0.9480 \\
PredNet & 12.5M & 8.6G & 31.85 & 90.01 & 0.9273 \\
PhyDNet & 3.1M & 15.3G & 20.35 & 61.47 & 0.9559 \\
PredRNN & 23.8M & 116.0G & 26.43 & 77.52 & 0.9411 \\
PredRNN++ & 38.6M & 171.7G & 14.07 & 48.91 & 0.9698 \\
MIM & 38.0M & 179.2G & 14.73 & 52.31 & 0.9678 \\
MAU & 4.5M & 17.8G & 22.25 & 67.96 & 0.9511 \\
E3D-LSTM & 51.0M & 298.9G & 24.07 & 77.49 & 0.9436 \\
PredRNN.V2 & 23.9M & 116.6G & 17.26 & 57.22 & 0.9624 \\
SimVP+IncepU & 58.0M & 19.4G & 21.15 & 64.15 & 0.9536 \\
TAU & 44.7M & 16.0G & 15.69 & 51.46 & 0.9721 \\
\midrule
STLight-L (Ours) & 32.9M & 32.9M &\textbf{ 14.77} & \textbf{47.17} & \textbf{0.9686} \\ 
\bottomrule
\end{tabular}
} \label{long_runs}

\label{tab:model_comparison}
\end{table}

\section{STLight method complexity}\label{sec:complexity}\label{parametercount}
To analyze the complexity of the STLight method, we'll break down its main components: Spatio-Temporal Patches, Patch Shuffle and Reassemble, and the Repeated STLMixer setup. By examining the number of parameters each part uses, we can gain insights into the method's design and its computational demands.

\begin{itemize}
    \item\textbf{Spatio-Temporal Patches} STLight encodes the input frames with a parameter count of $O\left((T C) \cdot d \cdot k_E^2\right)$ because it uses a single convolution, with input channels, output channels and kernel size equal to $T\cdot C$, $d$, $k_E$ respectively.

    \item \textbf{Patch Shuffle and Reassemble} While the patch shuffle layer doesn't have any learnable parameters, the patch reassemble is based on a single pointwise convolution with input channels, output channels and kernel size respectively equal to $d/p^2$, $T' \cdot C$, $k_D=1$. Therefore STLight decodes the processed signals with a parameter count of $O\left(d/p^2 \cdot (T' C) \cdot k_D^2 \right)$ = $O\left(d/p^2 \cdot (T' C) \right)$. 

    \item\textbf{Repeated STLMixer} The parameter count from our proposed STLMixer architecture is $O\left(\texttt{de} \cdot (d^2 + d \cdot k_{T_1}^2 + d\cdot k_{T_2}^2 )\right)$. In fact, each STLMixer is composed of two depthwise convolutions and one pointwise convolutions. Each convolution has the same input channels and output channel dimensions that are equal to $d$. Depthwise convolutions perform group convolution with $\textrm{group\_size} = d$, hence their parameter counts are $O(d \cdot k_{T_1}^2)$ and $O(d \cdot k_{T_2}^2)$. The pointwise convolution has \(1 \times 1\) kernel size, so the parameter count for this layer is $O(d^2)$. Summing the three terms and considering that they are \texttt{de} repeated blocks inside STLight, we obtain the aforementioned complexity.

\end{itemize}

\noindent In order to simplify the parameter count formulas, we consider the following assumptions: 

\begin{enumerate}
    \item  \( T \cdot C \) and \( T' \cdot C \) typically remain below 10, while \( d \) often exceeds 1000, hence \( T \cdot C \ll d \) and \( T' \cdot C \ll d \). Similarly \(k_{T_1}^2 \ll d\) and \(k_{T_2}^2 \ll d\).
    \item  $O \leq 2$  and $p \leq 2$, leading to $k_E = p\cdot \textrm{max}(1, O) \leq 4$ using the formula shown in Section 2.1.
\end{enumerate}
Therefore the parameter counts of our encoder and decoder blocks scale linearly with respect to $d$, while the parameters count of the repeated STLMixer blocks can be expressed as $O(\texttt{de} \cdot d^2)$.

\boldmath \section {Optimal $k_{T_1}$ and $k_{T_2}$} \unboldmath \label{sec:kernel_size}This section explores the optimal kernel sizes $k_{T_1}$ and $k_{T_2}$ for the two depthwise convolutional stages within the STLMixer block. While large values for $k_{T_1}$ and $k_{T_2}$ ensure good local context and wider receptive field, they also increase the model's parameter count, possibly leading to worse performances due to overfitting. Our experiments, detailed in Figure \ref{fig:kernel_ablation}, show that a small $k_{T_1}=3$ combined with a larger $k_{T_2} \in \{5,7\}$ achieves optimal performance while maintaining a lower parameter count. They also indicate that excessively large kernel sizes not only decrease efficiency but also lead to poorer performance.
Given the consistency of these findings across various hyperparameters configurations and scenarios, we decided to keep $k_{T_1}$ and $k_{T_2}$ constant during our evaluations.


\begin{table}
\centering
\pgfplotstabletypeset[%
     color cells={min=22.33,max=24.51,textcolor=black},
     /pgfplots/colormap={blackwhite}{rgb255=(82, 190, 128) color=(white) rgb255=(205, 97, 85)},
    /pgf/number format/fixed,
    /pgf/number format/precision=2,
    col sep=comma,
     every head row/.style={before row=\toprule,after row=\midrule},
     every last row/.style={after row=\bottomrule},
    columns/$k_{T_1} / k_{T_2}$/.style={reset styles,string type}%
]{
$k_{T_1} / k_{T_2}$, 3,5,7,9,11
3, 24.16, 22.48, 22.33, 22.33,  22.75 
5,24.01, 22.67, 22.45, 22.78, 23.03 
7,23.75, 22.83, 22.88, 23.00, 24.00
9, 23.45, 23.02, 23.10, 23.38, 23.62
11, 23.04, 23.14, 23.24, 23.65, 24.51
}

\caption{STLight MSE comparison for different values of $k_{T_1}$ and $k_{T_2}$.}
\label{fig:kernel_ablation}
\end{table}


\section{Weight initialization}\label{sec:weight_init}
Initial weight settings are crucial for how quickly and effectively a deep learning model learns. In our study, we compare three different initializations: 
\begin{itemize}
    \item \textbf{Kaiming-Uniform} Following PyTorch's default weight initialization, we use uniform Kaiming initialization (also known as He initialization) \cite{he2015delving} for all the model's convolutional blocks.
    \item \textbf{Kaiming-Normal} As reported in Listing \ref{lst:initialization}, we initialize convolutional layers using gaussian Kaiming initialization.
    \item \textbf{Hybrid} We initialize the patch reassemble layer using the uniform Kaiming initialization and we initialize all the other layers following the approach mentioned in the Kaiming-Normal initialization. We hypothesize that the last layer requires a different initialization because it rearranges the shuffled patches and, unlike other layers, it is not responsible for processing spatial-temporal correlations.

\end{itemize}

\begin{lstlisting}[language=Python, caption=Code for the Kaiming-Normal weights initialization., label={lst:initialization}]
def _init_weights(self, m):
    if isinstance(m, nn.Conv2d):
        nn.init.kaiming_normal_(
            m.weight, 
            mode='fan_out', 
            nonlinearity='relu'
        )
        if m.bias is not None:
            nn.init.constant_(m.bias, 0)
            
\end{lstlisting}
    
In Figure \ref{fig:weight_init} we emphasize the importance of carefully selecting initial weight settings to guarantee stable training and reach optimal accuracy, by evaluating the three different initialization schemes mentioned above. 

Our experiments show that the Kaiming-Uniform initialization requires several epochs to stabilize before beginning to converge, while the all Kaiming-Normal approach does not lead to stable training. 

The optimal strategy is the Hybrid initialization, which leads to stable training while not requiring a number of initial epochs to stabilize.

\input{plots/weight_init}


\section{Optimal Training Settings} \label{sec:optimal_settings}

In Tables \ref{training_hyper_ablation} and \ref{fig
}, we evaluate the impact of various training hyperparameters on the performance of STLight-L on the MMNIST dataset to identify the most effective training hyperparameter configuration. The learning rate is a critical hyperparameter that influences the model's ability to learn from data while maintaining stability. A low learning rate leads to slow convergence, while a higher learning rate may cause instability, preventing convergence to optimal results \cite{Goodfellow-et-al-2016}. The final div factor determines the minimum learning rate achieved at the end of a training cycle with the OneCycleLR learning rate scheduler, influencing convergence behavior and model performance. We examine the effects of varying the learning rate (LR) and the number of training epochs on key metrics such as MSE, MAE, and SSIM. Tables \ref{training_hyper_ablation} and \ref{fig
} show that the default OpenSTL hyperparameters are highly effective for STLight-L, with potential improvements of 0.5 in MSE when using a learning rate of 0.003 instead of 0.001. In fact, in the first section of the table, we observe that reducing the learning rate from 0.003 to 0.0003 results in increased MSE and MAE, indicating that overly small learning rates hinder model performance. Conversely, increasing the learning rate to 0.01 leads to worse results, confirming that the optimal learning rate lies near 0.003. Table \ref{training_hyper_ablation} also shows that extending the number of training epochs improves all accuracy metrics, confirming that longer training durations significantly enhance model performance.

\begin{table}[h]

\centering

\caption{Ablation Study on STLight-L with Fixed Hyperparameters (dim=1400, depth=16, kernel\_size\_1=3, kernel\_size\_2=7, patch\_size=2) and Varying Training Learning Rate, and the Number of Epochs}
 \resizebox{0.45\textwidth}{!}{
\begin{tabular}{cccccc}
\toprule
\textbf{LR} & \textbf{Final Div} & \textbf{Epoch} & \textbf{MSE} & \textbf{MAE} & \textbf{SSIM} \\
\midrule
\multicolumn{6}{c}{$\uparrow$ Varying Learning Rate (LR)} \\
\midrule
0.0003 & 10000 & 200 & 27.42 & 76.86 & 0.937 \\
0.001  & 10000 & 200 & 22.28 & 66.11 & 0.950 \\
0.003  & 10000 & 200 & 21.80 & 64.90 & 0.951 \\
0.01   & 10000 & 200 & 38.47 & 90.42 & 0.939 \\
\midrule
\multicolumn{6}{c}{$\uparrow$ Varying Number of Epochs (Epoch)} \\
\midrule
0.001  & 10000 & 500  & 18.88 & 58.59 & 0.957 \\
0.001  & 10000 & 1000 & 17.89 & 54.25 & 0.962 \\
0.001  & 10000 & 2000 & 14.77 & 47.17 & 0.969 \\
\bottomrule
\end{tabular}
} \label{training_hyper_ablation}

\end{table}

\begin{table}[!h]
\centering
\pgfplotstabletypeset[%
     color cells={min=21.80,max=38.47,textcolor=black},
     /pgfplots/colormap={blackwhite}{rgb255=(82, 190, 128) color=(white) rgb255=(205, 97, 85)},
    /pgf/number format/fixed,
    /pgf/number format/precision=2,
    col sep=comma,
     every head row/.style={before row=\toprule,after row=\midrule},
     every last row/.style={after row=\bottomrule},
    columns/FinalDivFactor$/$lr/.style={reset styles,string type}%
]{

FinalDivFactor$/$lr, 0.0003,0.001,0.003,0.01
1000, 26.04, 22.41, 22.01, 35.99
3000,27.14, 22.49, 21.85,  38.35 
10000,27.42, 22.33,  21.8, 38.47
}
\caption{STLight-33M MSE comparison for different values of Learning Rate (lr) and FinalDivFactor.}
\label{fig:lr_ablation}
\end{table}

\section{Order of the decoding operations}    \label{sec:decoding_operations}
We investigate the optimal procedure for decoding the tensor $Z''_{T} \in \mathbb{R}^{B \times d \times H/p \times W/p}$ into the desired tensor of the predicted frames $\mathcal{B}_{T'}'$. We evaluated two choices:

\begin{itemize}
    \item \textbf{Shuffle-Reassemble} We first perform patch shuffle, obtaining a tensor of shape $B \times d/p^2 \times H \times W$. Subsequently, the tensor is reassembled using a $1\times 1$ convolutional layer, with $d/p^2$ input channels and $T' \cdot C$ output channels.
    \item \textbf{Reassemble-Shuffle} We first perform patch rearrange, using a $1\times 1$ convolutional layer, with $d$ input channels and $T' \cdot C \cdot p^2$ output channels, obtaining an intermediate tensor of shape $B \times (T' \cdot C \cdot p^2)  \times  H/p \times W/p$. Subsequently, we perform patch shuffle on the intermediate tensor, obtaining $B \times (T' \cdot C) \times H \times W$.
\end{itemize}

Both choices are able to effectively decode the output stage. In Figure \ref{fig:refiner}, we train STLight for different configuration settings and we report the standard deviation of the differences in loss values between consecutive epochs (``dispersion'') to get a measure of how much variation we get in the loss reduction. The figure clearly illustrates that Shuffle-Reassemble notably curtails the feature dispersion. This adjustment leads to an uptick in model performance and promotes stability.

\begin{figure}[H]
    \centering
    \resizebox{0.48\textwidth}{!}{
        \begin{tikzpicture}
\begin{axis}[
    xlabel={Parameters (in millions)},
    ylabel={Dispersion},
    xmin=10, xmax=40,
    ymin=1, ymax=6,
    legend style={at={(0.255,0.88)},anchor=west},
    legend style={font=\footnotesize},
    ymajorgrids=true,
    grid style=dashed,
]
\addplot[scatter src=explicit symbolic, mark=triangle*,color=orange, line width=0.25mm]
    coordinates {
    (12.834,3.389)
    (14.243,1.687)
    (15.653,1.843)
    (18.472,1.683)
    (19.643,1.522)
    (21.805,2.179)
    (23.967,2.054)
    (26.129,4.635)
    (28.291,1.687)
    (30.966,3.713)
    (37.579,1.74)
};

\addplot[scatter src=explicit symbolic, mark=square*,color=teal, ultra thin,mark options={scale=1},line width=0.25mm]
    coordinates {
    (12.834,1.474)
    (14.243,1.498)
    (15.653,1.631)
    (18.472,1.583)
    (19.643,1.476)
    (21.805,1.461)
    (23.967,1.551)
    (26.129,1.554)
    (28.291,1.577)
    (30.966,1.553)
    (37.579,1.442)
};




\legend{Dispersion on Reassemble-Shuffle, Dispersion on Shuffle-Reassemble}
\end{axis}
\end{tikzpicture}
}
\caption{Dispersion vs Number of parameters on different decoding operations order}
\label{fig:refiner}
\end{figure}


\section{STLight Code}\label{sec:STLight_code}
We report the full implementation of the STLight model. We encourage readers to use STLight through our OpenSTL implementation which will be publicly available. 

\newcommand{\customfont}[1]{
  \fontsize{#1}{#1 * \real{1mm}}\selectfont
}

\lstset{
  basicstyle=\customfont{2.4mm}\ttfamily,
}

     \begin{figure*}[t!]
    \centering
    \begin{lstlisting}[language=Python] 
import torch
import torch.nn as nn

class Residual(nn.Module):
    def __init__(self, fn):
        super().__init__()
        self.fn = fn

    def forward(self, x):
        return self.fn(x) + x

def STLMixer(dim, K_1, k_2):
    return nn.Sequential(
        Residual(  # depthwise convolution block
            nn.Sequential(
                nn.Conv2d(dim, dim, K_1, groups=dim, padding="same"),
                nn.Conv2d(dim, dim, K_2, groups=dim, padding="same", 
                          dilation=3),
                nn.GELU(),
                nn.BatchNorm2d(dim),
            )
        ),
        nn.Sequential(  # pointwise convolution block
            nn.Conv2d(dim, dim, kernel_size=1), 
            nn.GELU(), nn.BatchNorm2d(dim)
        ),
    )

def PatchEncoder(in_layers, dim, patch_size, overlapping):
    return nn.Sequential(
        nn.Conv2d(in_layers, dim, 
                  kernel_size=patch_size * max(1, overlapping),
                  stride=patch_size, 
                  padding=max(0, overlapping - 1) * patch_size // 2
                  ),
        nn.BatchNorm2d(dim),
        nn.GELU(),
    )

class STLight(nn.Module):
    def __init__(
        self, in_layers, out_layers, dim, depth, patch_size, 
        overlapping, K_1, K_2
    ):
        super().__init__()
        self.out_layers = out_layers
        self.patch_encoder = PatchEncoder(in_layers, dim, patch_size, 
                                          overlapping)
        self.net = nn.ModuleList([STLMixer(dim, K_1, K_2) for _ in range(depth)])
        self.patch_reassemble = nn.Conv2d(dim // patch_size**2,
                                          out_layers, kernel_size= 1)
        self.up = nn.PixelShuffle(patch_size)
        self.patch_encoder.apply(self._init_weights)
        self.net.apply(self._init_weights)

    def forward(self, x):
        B, T, C, H, W = x.shape
        x = x.reshape(B, T * C, H, W)
        x = self.patch_encoder(x)

        for i, block in enumerate(self.net):
            if i == len(self.net) // 3:
                x1 = x
            if i == 2 * len(self.net) // 3:
                x = x + x1
            x = block(x)
        x = self.up(x)
        x = self.patch_reassemble(x)
        return x.reshape(B, self.out_layers // C, C, H, W)

\end{lstlisting}

    \caption{Full STLight implementation}
    \label{fig:enter-label}
\end{figure*}

\clearpage

%% file: plots/weight_init.tex
\begin{figure}[H]
\centering
    \resizebox{0.48\textwidth}{!}{
\begin{tikzpicture}[font=\Large]
    \begin{axis}[
      xlabel={Epochs},
      ylabel={MSE},
      xmin=0, xmax=200,
      ymin=0, ymax=500,
      ymajorgrids=true,
      grid style=dashed,
      ymode=log,
      log basis y=2,
      legend pos=north east,
      width=0.9\textwidth,
    ]

    \addplot[name path=all_custom_down,color=green!70, line width=0.25mm] coordinates {(0,358.85)(1,229.549)(2,191.994)(3,172.729)(4,162.073)(5,155.16)(6,151.674)(7,146.285)(8,144.081)(9,138.731)(10,133.994)(11,131.285)(12,125.098)(13,122.123)(14,119.458)(15,113.849)(16,113.983)(17,107.649)(18,113.207)(19,120.449)(20,141.766)(21,163.738)(22,145.012)(23,271.077)(24,358.512)(25,152.795)(26,120.765)(27,97.69)(28,147.585)(29,123.683)(30,78.451)(31,65.329)(32,74.782)(33,67.89)(34,67.449)(35,64.132)(36,60.652)(37,60.899)(38,58.98)(39,57.557)(40,56.464)(41,54.601)(42,54.721)(43,52.664)(44,52.431)(45,51.284)(46,51.131)(47,49.917)(48,50.95)(49,50.235)(50,48.189)(51,47.968)(52,47.348)(53,47.878)(54,46.324)(55,45.567)(56,46.089)(57,46.155)(58,43.724)(59,44.517)(60,44.438)(61,43.491)(62,42.301)(63,43.415)(64,42.447)(65,41.933)(66,42.399)(67,41.385)(68,41.013)(69,40.749)(70,40.767)(71,40.624)(72,39.617)(73,39.043)(74,39.796)(75,38.985)(76,38.887)(77,38.57)(78,39.076)(79,38.261)(80,37.235)(81,38.948)(82,37.532)(83,37.19)(84,37.39)(85,35.869)(86,35.945)(87,36.091)(88,36.233)(89,34.382)(90,36.807)(91,35.059)(92,35.164)(93,35.309)(94,34.942)(95,35.33)(96,33.724)(97,34.088)(98,34.071)(99,33.432)(100,34.184)(101,33.626)(102,33.365)(103,32.923)(104,32.87)(105,33.277)(106,32.631)(107,33.248)(108,32.131)(109,31.712)(110,32.18)(111,31.395)(112,31.34)(113,31.214)(114,31.8)(115,31.401)(116,31.431)(117,30.864)(118,30.747)(119,31.131)(120,30.28)(121,30.964)(122,30.434)(123,30.154)(124,30.229)(125,29.713)(126,29.641)(127,29.359)(128,29.436)(129,29.344)(130,29.237)(131,29.189)(132,28.981)(133,29.07)(134,28.825)(135,28.669)(136,28.813)(137,28.243)(138,28.559)(139,28.258)(140,27.953)(141,28.033)(142,27.752)(143,27.687)(144,27.664)(145,27.702)(146,27.42)(147,27.511)(148,27.533)(149,27.438)(150,27.196)(151,27.099)(152,26.88)(153,26.837)(154,26.865)(155,26.535)(156,26.699)(157,26.456)(158,26.47)(159,26.286)(160,26.324)(161,26.303)(162,26.186)(163,26.123)(164,26.046)(165,25.99)(166,25.89)(167,25.747)(168,25.844)(169,25.685)(170,25.585)(171,25.591)(172,25.501)(173,25.507)(174,25.456)(175,25.442)(176,25.427)(177,25.335)(178,25.341)(179,25.25)(180,25.178)(181,25.216)(182,25.232)(183,25.155)(184,25.109)(185,25.136)(186,25.079)(187,25.066)(188,25.041)(189,25.043)(190,25.009)(191,25.001)(192,24.986)(193,25.007)(194,24.971)(195,24.987)(196,24.958)(197,24.981)(198,24.983)(199,24.985)};
    
    \addplot[name path=all_custom_up,color=green!70, line width=0.25mm] coordinates {(0,368.905)(1,239.026)(2,199.092)(3,180.608)(4,169.981)(5,158.94)(6,153.905)(7,150.164)(8,148.172)(9,142.681)(10,137.176)(11,134.967)(12,128.658)(13,124.565)(14,121.884)(15,121.72)(16,115.338)(17,113.809)(18,128.497)(19,124.094)(20,155.493)(21,211.331)(22,210.809)(23,331.133)(24,436.679)(25,434.571)(26,439.568)(27,514.828)(28,287.211)(29,374.793)(30,336.464)(31,206.847)(32,199.22)(33,141.464)(34,110.66)(35,80.733)(36,68.548)(37,65.7)(38,62.654)(39,61.297)(40,58.52)(41,65.98)(42,58.409)(43,54.783)(44,54.694)(45,53.806)(46,53.12)(47,53.799)(48,52.166)(49,52.389)(50,53.38)(51,49.579)(52,60.886)(53,48.422)(54,49.776)(55,47.059)(56,47.19)(57,47.372)(58,45.742)(59,44.81)(60,45.893)(61,65.463)(62,47.068)(63,276.341)(64,47.15)(65,44.798)(66,44.722)(67,42.568)(68,41.614)(69,41.891)(70,42.941)(71,42.242)(72,41.573)(73,40.65)(74,41)(75,41.095)(76,62.735)(77,49.117)(78,43.339)(79,50.28)(80,47.065)(81,44.164)(82,47.769)(83,39.04)(84,40.058)(85,43.674)(86,39.737)(87,42.419)(88,40.579)(89,37.6)(90,40.757)(91,43.502)(92,43.082)(93,37.1)(94,44.027)(95,37.059)(96,35.086)(97,36.129)(98,37.262)(99,35.678)(100,34.983)(101,35.693)(102,45.513)(103,50.33)(104,89.062)(105,62.717)(106,61.931)(107,61.008)(108,34.909)(109,41.301)(110,50.045)(111,46.732)(112,41.073)(113,46.762)(114,35.843)(115,36.409)(116,37.2)(117,40.177)(118,39.405)(119,37.885)(120,42.065)(121,34.315)(122,36.101)(123,35.311)(124,36.43)(125,41.289)(126,44.405)(127,41.485)(128,37.013)(129,46.412)(130,45.018)(131,38.786)(132,31.38)(133,30.718)(134,35.266)(135,31.398)(136,61.692)(137,51.931)(138,39.078)(139,43.399)(140,40.052)(141,41.929)(142,34.151)(143,49.279)(144,38.609)(145,40.719)(146,40.099)(147,47.902)(148,30.736)(149,42.339)(150,32.53)(151,30.273)(152,29.092)(153,30.454)(154,29.345)(155,30.537)(156,28.645)(157,27.831)(158,27.747)(159,28.175)(160,28.207)(161,34.516)(162,30.84)(163,29.223)(164,28.522)(165,28.273)(166,29.325)(167,27.596)(168,32.299)(169,28.948)(170,28.061)(171,28.562)(172,28.474)(173,28.588)(174,30.205)(175,31.865)(176,31.296)(177,31.08)(178,30.842)(179,30.051)(180,28.501)(181,26.953)(182,31.262)(183,27.747)(184,33.386)(185,34.615)(186,27.268)(187,30.536)(188,26.798)(189,30.034)(190,31.095)(191,26.595)(192,28.257)(193,30.519)(194,32.145)(195,33.53)(196,29.476)(197,31.078)(198,30.931)(199,28.374)};
    \addplot[green!50,fill opacity=0.2, line width=0.25mm] fill between[of=all_custom_down and all_custom_up];
    
    \addplot[name path=all_standard_down,color=blue!70, line width=0.25mm] coordinates {(0,139.258)(1,124.853)(2,116.77)(3,110.945)(4,105.509)(5,102.022)(6,97.083)(7,94.153)(8,91.327)(9,88.712)(10,87.081)(11,83.535)(12,80.817)(13,81.317)(14,79.505)(15,74.833)(16,74.624)(17,71.533)(18,69.864)(19,69.889)(20,69.112)(21,68.124)(22,66.369)(23,68.082)(24,64.597)(25,64.695)(26,63.231)(27,62.957)(28,59.53)(29,59.553)(30,60.019)(31,60.301)(32,58.444)(33,58.953)(34,58.196)(35,56.283)(36,56.594)(37,57.968)(38,54.128)(39,56.565)(40,56.321)(41,53.911)(42,52.589)(43,53.232)(44,52.068)(45,50.754)(46,51.421)(47,51.77)(48,52.014)(49,49.255)(50,48.9)(51,47.422)(52,47.409)(53,48.358)(54,48.987)(55,46.435)(56,48.515)(57,46.94)(58,46.073)(59,45.042)(60,44.369)(61,44.287)(62,44.305)(63,43.99)(64,42.904)(65,43.606)(66,44.038)(67,43.812)(68,45.019)(69,41.554)(70,40.508)(71,42.438)(72,41.566)(73,42.824)(74,42.778)(75,46.051)(76,42.113)(77,41.17)(78,45.794)(79,40.963)(80,40.458)(81,41.961)(82,52.816)(83,41.416)(84,46.496)(85,38.894)(86,41.975)(87,52.685)(88,41.46)(89,39.313)(90,39.025)(91,41.294)(92,42.475)(93,50.631)(94,37.462)(95,55.019)(96,38.679)(97,45.731)(98,40.327)(99,48.433)(100,43.538)(101,40.033)(102,37.504)(103,38.596)(104,41.328)(105,37.411)(106,37.561)(107,35.618)(108,36.365)(109,35.504)(110,33.473)(111,34.124)(112,35.448)(113,35.19)(114,38.318)(115,34.719)(116,33.735)(117,33.369)(118,36.962)(119,33.961)(120,32.969)(121,33.886)(122,37.878)(123,32.099)(124,33.881)(125,33.941)(126,33.313)(127,34.987)(128,34.836)(129,34.578)(130,33.584)(131,34.952)(132,33.431)(133,32.683)(134,31.952)(135,31.777)(136,37.913)(137,34.459)(138,33.878)(139,33.399)(140,32.958)(141,31.715)(142,34.56)(143,34.707)(144,31.851)(145,31.592)(146,31.318)(147,33.615)(148,30.63)(149,36.152)(150,30.267)(151,32.591)(152,30.034)(153,31.199)(154,30.963)(155,34.768)(156,30.059)(157,30.279)(158,32.315)(159,33.099)(160,33.388)(161,33.578)(162,31.984)(163,29.723)(164,41.142)(165,33.312)(166,29.693)(167,33.153)(168,29.86)(169,33.298)(170,32.166)(171,28.662)(172,33.453)(173,30.006)(174,31.73)(175,29.846)(176,28.506)(177,30.938)(178,33.93)(179,31.02)(180,29.774)(181,31.319)(182,28.871)(183,30.745)(184,30.168)(185,32.357)(186,40.642)(187,33.563)(188,40.448)(189,28.665)(190,29.871)(191,30.191)(192,34.109)(193,32.508)(194,33.84)(195,33.407)(196,27.795)(197,33.351)(198,32.143)(199,35.108)};
    
    \addplot[name path=all_standard_up,color=blue!70, line width=0.25mm] coordinates {(0,143.469)(1,128.268)(2,120.901)(3,114.814)(4,111.354)(5,103.101)(6,99.143)(7,96.22)(8,94.776)(9,94.009)(10,91.129)(11,100.926)(12,87.201)(13,90.072)(14,95.579)(15,78.313)(16,90.582)(17,76.841)(18,87.759)(19,78.195)(20,71.433)(21,69.78)(22,82.017)(23,77.348)(24,66.513)(25,66.327)(26,72.935)(27,70.811)(28,64.855)(29,72.043)(30,73.526)(31,90.029)(32,69.154)(33,60.427)(34,65.158)(35,59.486)(36,60.828)(37,60.946)(38,64.278)(39,67.938)(40,74.891)(41,66.147)(42,56.878)(43,55.192)(44,67.657)(45,75.285)(46,59.763)(47,53.352)(48,59.22)(49,56.667)(50,56.057)(51,57.56)(52,61.202)(53,60.161)(54,78.153)(55,56.78)(56,64.319)(57,63.338)(58,58.011)(59,55.143)(60,52.127)(61,51.054)(62,53.257)(63,48.468)(64,62.127)(65,52.432)(66,52.45)(67,46.148)(68,51.103)(69,59.426)(70,53.814)(71,60.828)(72,53.386)(73,62.646)(74,47.045)(75,65.349)(76,61.417)(77,44.994)(78,51.673)(79,49.125)(80,42.577)(81,48.317)(82,84.082)(83,62.493)(84,60.94)(85,53.954)(86,75.778)(87,71.015)(88,58.45)(89,58.52)(90,43.725)(91,46.623)(92,52.87)(93,58.082)(94,61.355)(95,57.888)(96,51.421)(97,50.071)(98,57.84)(99,62.167)(100,60.35)(101,49.235)(102,52.563)(103,57.634)(104,63.951)(105,62.71)(106,38.96)(107,52.958)(108,44.039)(109,50.515)(110,52.408)(111,56.383)(112,47.493)(113,57.746)(114,71.661)(115,68.459)(116,52.576)(117,45.225)(118,60.845)(119,50.023)(120,80.301)(121,59.632)(122,78.366)(123,77.176)(124,42.092)(125,77.109)(126,97.457)(127,55.811)(128,62.445)(129,92.261)(130,49.406)(131,81.336)(132,74.023)(133,82.433)(134,57.223)(135,73.869)(136,103.748)(137,49.601)(138,50.696)(139,59.177)(140,85.469)(141,60.829)(142,90.973)(143,76.983)(144,50.02)(145,72.252)(146,68.462)(147,47.834)(148,76.188)(149,64.312)(150,68.297)(151,69.465)(152,75.634)(153,69.561)(154,68.45)(155,66.632)(156,56.639)(157,54.158)(158,60.112)(159,71.735)(160,58.664)(161,53.26)(162,91.134)(163,51.367)(164,74.686)(165,83.656)(166,73.798)(167,78.702)(168,79.429)(169,81.037)(170,67.132)(171,69.147)(172,67.187)(173,92.193)(174,50.327)(175,62.737)(176,58.922)(177,75.587)(178,68.525)(179,103.327)(180,69.428)(181,102.477)(182,59.978)(183,123.834)(184,69.309)(185,52.157)(186,60.685)(187,57.495)(188,68.752)(189,99.278)(190,93.355)(191,59.35)(192,94.998)(193,60.903)(194,57.132)(195,102.026)(196,82.493)(197,65.636)(198,81.963)(199,79.228)};
    \addplot[blue!50,fill opacity=0.2] fill between[of=all_standard_down and all_standard_up];

    \addplot[name path=hybrid_down,color=red!70, line width=0.25mm] coordinates {(0,137.266)(1,125.075)(2,117.952)(3,111.897)(4,106.49)(5,103.071)(6,98.506)(7,95.473)(8,92.493)(9,89.103)(10,85.657)(11,84.591)(12,80.516)(13,80.146)(14,76.891)(15,75.458)(16,73.78)(17,71.188)(18,69.197)(19,68.229)(20,66.207)(21,65.282)(22,64.379)(23,62.219)(24,60.633)(25,60.222)(26,58.189)(27,57.761)(28,57.055)(29,56.362)(30,54.657)(31,54.557)(32,54.331)(33,52.192)(34,52.277)(35,52.473)(36,51.326)(37,50.958)(38,49.727)(39,49.026)(40,48.542)(41,47.767)(42,46.356)(43,46.64)(44,46.093)(45,45.968)(46,44.877)(47,44.47)(48,43.736)(49,43.486)(50,43.551)(51,43.833)(52,42.589)(53,42.46)(54,42.069)(55,41.844)(56,41.006)(57,40.846)(58,40.42)(59,40.195)(60,39.929)(61,38.289)(62,38.504)(63,38.495)(64,38.214)(65,38.681)(66,37.904)(67,37.865)(68,37.447)(69,37.013)(70,36.986)(71,36.29)(72,36.009)(73,36.337)(74,35.365)(75,35.972)(76,35.524)(77,35.031)(78,34.744)(79,34.746)(80,34.686)(81,34.36)(82,34.225)(83,33.783)(84,34.126)(85,33.842)(86,33.702)(87,33.751)(88,32.937)(89,32.631)(90,32.56)(91,32.674)(92,32.321)(93,32.179)(94,32.078)(95,32.146)(96,31.595)(97,31.87)(98,31.476)(99,31.381)(100,31.218)(101,31.076)(102,30.965)(103,31.153)(104,31.028)(105,30.646)(106,30.452)(107,30.41)(108,30.321)(109,30.188)(110,29.986)(111,29.8)(112,29.846)(113,29.953)(114,29.372)(115,29.82)(116,28.916)(117,29.134)(118,28.932)(119,28.951)(120,28.792)(121,28.792)(122,28.537)(123,28.545)(124,28.186)(125,28.288)(126,28.255)(127,27.832)(128,28.016)(129,27.788)(130,27.605)(131,27.675)(132,27.461)(133,27.171)(134,27.151)(135,27.011)(136,27.159)(137,26.901)(138,26.846)(139,26.684)(140,26.713)(141,26.725)(142,26.608)(143,26.416)(144,26.413)(145,26.292)(146,26.029)(147,25.998)(148,26.064)(149,26.077)(150,25.805)(151,25.515)(152,25.7)(153,25.581)(154,25.434)(155,25.385)(156,25.256)(157,25.264)(158,25.269)(159,25.166)(160,25.081)(161,25.002)(162,24.96)(163,24.96)(164,24.818)(165,24.752)(166,24.58)(167,24.583)(168,24.566)(169,24.454)(170,24.422)(171,24.398)(172,24.406)(173,24.288)(174,24.256)(175,24.206)(176,24.194)(177,24.161)(178,24.117)(179,24.051)(180,24.03)(181,23.988)(182,24.028)(183,23.97)(184,23.913)(185,23.907)(186,23.87)(187,23.849)(188,23.849)(189,23.869)(190,23.824)(191,23.838)(192,23.854)(193,23.83)(194,23.814)(195,23.795)(196,23.801)(197,23.797)(198,23.801)(199,23.802)};
    
    \addplot[name path=hybrid_up,color=red!70, line width=0.25mm] coordinates {(0,138.058)(1,126.612)(2,120.415)(3,115.053)(4,110.533)(5,105.946)(6,104.853)(7,98.702)(8,95.495)(9,92.158)(10,88.612)(11,85.908)(12,83.302)(13,81.884)(14,78.268)(15,76.618)(16,74.794)(17,73.192)(18,70.537)(19,70.387)(20,68.132)(21,65.669)(22,66.349)(23,64.064)(24,63.427)(25,61.262)(26,59.723)(27,58.493)(28,57.265)(29,56.66)(30,56.269)(31,55.616)(32,55.05)(33,57.383)(34,53.205)(35,52.745)(36,53.057)(37,52.004)(38,52.644)(39,49.984)(40,53.865)(41,49.031)(42,47.906)(43,47.069)(44,48.717)(45,46.601)(46,47.422)(47,48.492)(48,47.609)(49,44.72)(50,44.826)(51,45.338)(52,43.483)(53,43.564)(54,42.171)(55,42.91)(56,42.945)(57,42.213)(58,42.35)(59,42.112)(60,41.119)(61,39.44)(62,39.591)(63,39.564)(64,39.159)(65,38.963)(66,38.944)(67,38.629)(68,38.235)(69,37.583)(70,38.391)(71,36.734)(72,37.215)(73,36.814)(74,35.938)(75,37.17)(76,36.342)(77,35.38)(78,35.793)(79,35.398)(80,38.016)(81,34.969)(82,35.87)(83,34.791)(84,34.566)(85,34.268)(86,34.95)(87,33.957)(88,34.152)(89,32.712)(90,32.853)(91,34.562)(92,33.239)(93,32.585)(94,32.46)(95,32.835)(96,32.137)(97,32.19)(98,32.014)(99,31.68)(100,31.62)(101,33.484)(102,31.572)(103,31.548)(104,31.388)(105,31.516)(106,31.499)(107,30.631)(108,30.809)(109,30.474)(110,31.076)(111,31.56)(112,31.538)(113,30.008)(114,29.719)(115,29.867)(116,31.451)(117,29.728)(118,29.374)(119,29.638)(120,29.139)(121,29.179)(122,29.067)(123,28.714)(124,28.963)(125,29.142)(126,29.294)(127,28.646)(128,28.135)(129,27.882)(130,28.212)(131,28.163)(132,27.997)(133,27.892)(134,27.811)(135,28.271)(136,27.664)(137,27.488)(138,27.26)(139,27.159)(140,27.203)(141,26.967)(142,27.229)(143,26.896)(144,26.847)(145,26.769)(146,26.654)(147,27.17)(148,26.654)(149,26.522)(150,26.484)(151,26.58)(152,26.263)(153,26.087)(154,25.964)(155,26.268)(156,25.839)(157,25.989)(158,25.945)(159,25.867)(160,25.805)(161,25.538)(162,26.017)(163,25.481)(164,25.395)(165,25.418)(166,25.56)(167,25.551)(168,25.28)(169,25.288)(170,25.132)(171,25.057)(172,25.012)(173,25.641)(174,25.012)(175,24.944)(176,24.987)(177,25.402)(178,24.856)(179,24.999)(180,24.781)(181,24.785)(182,24.722)(183,24.71)(184,24.866)(185,24.975)(186,24.643)(187,24.817)(188,24.829)(189,25.119)(190,24.624)(191,24.962)(192,24.633)(193,24.792)(194,24.582)(195,24.656)(196,24.747)(197,24.728)(198,24.941)(199,24.517)};
    \addplot[red!50,fill opacity=0.2] fill between[of=hybrid_down and hybrid_up];
    \legend{Normal Initialization, , , Standard Initialization, , , Hybrid Initialization, , }
100    
    \end{axis}
\end{tikzpicture}
}
\caption{Learning curve comparison for three different weights initializations. For each of them, we report the mimimum and maximum boundaries of the MSE out of 5 runs.}
\label{fig:weight_init}
\end{figure}